\documentclass[11pt, a4paper]{article}
\usepackage{amsmath,amssymb,amsfonts,amsthm,enumerate,color,ifthen}
\usepackage{xargs}

\usepackage{algorithm}
\usepackage{algpseudocode}
\usepackage{subcaption}

\usepackage{float}
\usepackage{graphicx}

\usepackage{aliascnt,bbm}

\newtheorem{rem}{Remark}[section]

\def\PE{\mathbb{E}}
\def\rset{\mathbb{R}}
\def\T{{ \mathrm{\scriptscriptstyle T} }}

\def\rset{\mathbb R}
\def\zset{\mathbb Z}

\def\eqsp{\;}

\newcommand{\pscal}[2]{\left\langle#1,#2\right\rangle}

\newcommand{\eqdef}{\ensuremath{\stackrel{\mathrm{def}}{=}}}

\def\Xset{\mathcal{X}} 
\def\cB{\mathsf{B}} 

\def\M{\mathcal{M}}

\newcommandx\sequence[3][2=t,3=\zset]
{\ifthenelse{\equal{#3}{}}{\ensuremath{\{ #1_{#2}\}}}{\ensuremath{\{ #1_{#2}, \eqsp #2 \in #3 \}}}}

\def\PP{\mathbb{P}} 
\newcommand{\CPP}[3][]
{\ifthenelse{\equal{#1}{}}{{\mathbb P}\left(\left. #2 \, \right| #3 \right)}{{\mathbb P}_{#1}\left(\left. #2 \, \right | #3 \right)}}
\def\PE{\mathbb{E}} 
\newcommand{\CPE}[3][]
{\ifthenelse{\equal{#1}{}}{{\mathbb E}\left[\left. #2 \, \right| #3 \right]}{{\mathbb E}_{#1}\left[\left. #2 \, \right | #3 \right]}}

\def\tv{\mathrm{tv}}

\def\Cset{\mathcal{C}} 

\def\I{\textsf{I}}

\def\r{\textsf{r}}

\usepackage{bm}

\theoremstyle{plain}
\newtheorem{theorem}{Theorem}
\newtheorem{assumption}{H\hspace{-3pt}}

\newaliascnt{proposition}{theorem}
\newtheorem{proposition}[proposition]{Proposition}
\aliascntresetthe{proposition}

\newaliascnt{lemma}{theorem}
\newtheorem{lemma}[lemma]{Lemma}
\aliascntresetthe{lemma}
\newaliascnt{corollary}{theorem}

\aliascntresetthe{corollary}

\theoremstyle{definition}
\newaliascnt{definition}{theorem}
\newtheorem{definition}[definition]{Definition}
\aliascntresetthe{definition}

\newaliascnt{remark}{theorem}
\newtheorem{remark}[remark]{Remark}
\aliascntresetthe{remark}

\newaliascnt{example}{theorem}

\aliascntresetthe{example}

\def\rmd{\mathrm{d}}

\def\1{\mathbbm{1}}

\usepackage{verbatim}
\usepackage{nicefrac}
\usepackage{epstopdf}
\usepackage{color}
\usepackage{textcomp} 
\usepackage{rotating}
\usepackage{paralist} 
\usepackage{booktabs}
\usepackage{bm}
\usepackage{url} 
\usepackage[toc,page]{appendix}
\usepackage{epsfig}
\usepackage{bbm}
\usepackage{indentfirst}
\usepackage{authblk}

\newcommand{\bls}[1]{\renewcommand{\baselinestretch}{0.9}\footnotesize\normalsize}

\begin{document}
	\date{}
\title{\bf Minimax quasi-Bayesian estimation in sparse canonical correlation analysis via a Rayleigh quotient function}
\author{Qiuyun Zhu\hspace{.0cm} and 
	Yves Atchad\'e\thanks{
		The authors gratefully acknowledge \textit{NSF grant DMS 2015485.} The authors are grateful to Roger Zoh for very helpful discussions.
	}
	\\
	Department of Mathematics and Statistics,  Boston University}
\maketitle
\begin{abstract}
	Canonical correlation analysis (CCA) is a popular statistical technique for exploring relationships between datasets. In recent years, the estimation of sparse canonical vectors has emerged  as an important but challenging variant of the CCA problem, with widespread applications.   
	Unfortunately, existing rate-optimal estimators  for sparse canonical vectors have high computational cost. We propose a quasi-Bayesian estimation procedure that not only achieves the minimax estimation rate, but also is easy to compute by Markov Chain Monte Carlo (MCMC). The method builds on (\cite{tan2018sparse}) and uses a re-scaled Rayleigh quotient function as the quasi-log-likelihood. However, unlike  (\cite{tan2018sparse}), we adopt a Bayesian framework that combines this quasi-log-likelihood with a spike-and-slab prior to regularize the inference and promote sparsity. We investigate the empirical behavior of the proposed method on both continuous and truncated data, and we demonstrate that it outperforms several state-of-the-art methods.
	As an application, we use the proposed methodology to maximally correlate clinical variables and proteomic data for better understanding the Covid-19 disease.
\end{abstract}

\section{Introduction}
\label{sec:intro}
Canonical correlation analysis (CCA) is a statistical technique --dating back at least to  \cite{hotelling1936relations} -- that is used to maximally correlate multiple datasets for joint analysis. The technique has become a fundamental tool  in biomedical research where technological advances have made it possible to observe fundamental biological phenomena from multiple viewpoints --- the so-called multi-omic datasets (\cite{witten2009extensions,mo2017fully,rappoport2018multi}). Over the past two decades, limited sample size and growing dimensionality in these datasets, and the search for meaningful biological interpretations, have led to the development of sparse CCA (\cite{wiesel:etal:08,witten2009extensions, parkhomenko:etal:09, waaijenborg:zwinderman:09, hardoon:etal:11}),  where a sparsity assumption is imposed on the canonical vectors.

Statistically optimal estimation of sparse CCA has been recently considered in the literature. (\cite{gao2015}) derived the minimax rate of estimation of sparse CCA, and proposed a two-stage estimation procedure that achieves the rate. (\cite{tan2018sparse}) uses a generalized Rayleigh quotient approach to propose a two-stage estimator that also achieves the minimax rate. 
These two  rate-optimal estimation procedures share the same limitation, that is, high computational cost. 
Specifically, in both approaches, each iteration of the first-stage optimization problem  has a computational cost of $O(p^3)$, where $p$ is the joint number of variables in the datasets. Furthermore, the two-stage nature of these estimators can also be a problem in practice, since it can be hard to set the required stopping criterion of the first-stage solver  that guarantees a good behavior of the final estimator.  

We address these issues by proposing a conceptually simple, yet rate-optimal quasi-Bayesian estimator for sparse CCA. 
More specifically, building on (\cite{tan2018sparse}), we propose a quasi-Bayesian approach that  employs a re-scaled version of the Rayleigh quotient function as the quasi-log-likelihood  together with a spike and slab prior to obtain a quasi-posterior distribution. The method is agnostic to the covariance matrix estimators used in constructing the Rayleigh quotient function. For example, we observe in our experiments that both the sample covariance matrix estimator and the Kendall's-tau-based covariance matrix estimator (\cite{yoon:etal:18}) can be used to construct the Rayleigh quotient function, and these matrices are allowed to be singular. Although we do not pursue this here, one can straightforwardly extend our method to solve other generalized eigenvalue problems in the same spirit as (\cite{tan2018sparse}). In fact, at a high level, our method can be viewed as an improved version of simulated annealing (\cite{kirkpatrick:etal:83,bertsimas:tsitsiklis:93}) for minimizing the Rayleigh quotient under a sparsity constraint. As such, it can be easily extended to tackle other similarly challenging non-convex statistical optimization problems with sparsity constraints.

We analyze the proposed estimator and derive its convergence rate  (see Theorem \ref{thm:1}).  In the particular case where sample covariance matrices are used to estimate the Rayleigh quotient, we show that the estimator achieves the minimax rate for sparse CCA estimation, under some modest sample size conditions.

We propose a Markov Chain Monte Carlo algorithm based on simulated tempering to sample from the quasi-posterior distribution, and compute the estimator.  At stationarity, the proposed algorithm  has a per-iteration cost of $O(\bar s^2 p)$, where $\bar s$ is the underlying sparsity level of the posterior distribution. In all our numerical experiments, we have observed that $\bar s$ is of the same order as $s_\star$, namely the true sparsity level of the principal canonical vectors, leading to a very small percentage of false-positives. Furthermore,  we show empirically that for sufficiently large sample size, the mixing time of the algorithm scales linearly in $p$. As a result,  our estimator has a much lower computational cost than the \textsf{Rifle} estimator in (\cite{tan2018sparse}). We also compare our method with the popular \textsf{mixedCCA} estimator in (\cite{yoon:etal:18}). The results show that  although our method is computationally slower than \textsf{mixedCCA},  it produces statistically better estimates. We note that the estimation rate of \textsf{mixedCCA} is currently unknown.

The paper is organized as follows. In Section \ref{sec:meth} we introduce our estimation procedure and derive its convergence rate. In Section \ref{sec:mcmc} we detail a simulated tempering algorithm to sample from the resulting quasi-posterior distribution. In Section \ref{sec:num}, we study the behavior of the proposed method on both continuous and truncated data, and compare it with other methods. In Section~\ref{sec:covid}, we apply the method to a case study,  where one aims to correlate clinical and proteomic data from Covid-19 patients, for a better understanding of the disease. Our analysis identifies that Alpha-1-acid glycoprotein 1 (AGP 1) plays an important role in  the progression of Covid-19 into a severe illness.

A \textsf{Python} code is available from \url{https://github.com/rachelwho/Sparse-CCA}.

\section{Quasi-Bayesian sparse CCA using a Rayleigh quotient function}\label{sec:meth}

Let $(X,Y)\in\rset^{p_x}\times \rset^{p_y}$ be a pair of high-dimensional zero-mean  random vectors with joint distribution $f$ and covariance matrices $\Sigma_x\eqdef\PE(XX^\T)$, $\Sigma_y\eqdef \PE(YY^\T)$ and $\Sigma_{xy}\eqdef \PE(XY^\T)$.  Let $(v_{x\star},v_{y\star})\in\rset^{p_x}\times \rset^{p_y}$  be a pair of principal canonical vectors of $f$, that is, a vector 
pair that solves the following optimization problem: 
\begin{equation}\label{cca:probl:0}
	\max_{v_x \in\rset^{p_x},\;v_y\in\rset^{p_y}}\; v_x^T \Sigma_{xy}v_y\;\;\mbox{ s.t.}\;\;\;\; v_x^\T\Sigma_{x}v_x = v_y^\T \Sigma_{y}v_y = 1.
\end{equation}
Since we are only interested in  the directions of $v^\T_{x\star}$ and $v^\T_{y\star}$, we set $\theta_\star\eqdef \frac{(v^\T_{x\star},v^\T_{y\star})^\T}{\|(v^\T_{x\star},v^\T_{y\star})^\T\|_2}$ (so that $\|\theta_\star\|_2=1$) to be our main parameter of interest. The parameter $\theta_\star$  is  identifiable only up to a change of sign, and hence, we shall focus on the estimation of the related projector $\theta_\star\theta_\star^\T$. Let us define $p\eqdef p_x+p_y$, and the matrices
\begin{equation}
	A \eqdef \left[\begin{array}{cc} 0 & \Sigma_{xy}\\ \Sigma_{xy}^\T & 0 \end{array}\right],\;\;  \quad  B \eqdef \left[\begin{array}{cc} \Sigma_{x} & 0\\ 0 & \Sigma_{y}\end{array}\right] \;\;\; \mbox{ and } \;\;\;\Sigma \eqdef A + B = \left[\begin{array}{cc} \Sigma_{x} & \Sigma_{xy}\\ \Sigma_{xy}^\T & \Sigma_{y} \end{array}\right].\label{cca:A_B}
\end{equation}
Using simple arguments, we notice that the problem in (\ref{cca:probl:0}) is equivalent to the following generalized eigenvalue problem (GEP): 
\begin{equation}\label{cca:probl:1}
	\max_{\theta = (v_x^\T,v_y^\T)^\T\in\rset^{ p }}\; \theta^\T A \theta\;\; \mbox{ s.t.}\;\;\;\; \theta^\T B\theta= 2.
\end{equation}
Clearly, finding a solution of~\eqref{cca:probl:1}  is	equivalent to finding  a solution of 
\begin{equation}
	\max_{\theta = (v_x^\T,v_y^\T)^\T\in\rset^{p}}\textsf{R}(\theta) \eqdef \frac{\theta^\T A\theta}{\theta^\T B\theta},\label{cca:R}
\end{equation}
where we convene that $0/0=0$. The objective function $\textsf{R}(\cdot)$ in~\eqref{cca:R} is known as the {\em (generalized) Rayleigh quotient} of $A$ and $B$. 
The reformulation in \eqref{cca:R} suggests a way to 
estimate the sparse canonical vectors by directly targeting the Rayleigh quotient, and this idea was first proposed in (\cite{tan2018sparse}). Note that solving~\eqref{cca:R}  requires specifying matrices $A$ and $B$, which are typically unknown in practice. Instead, given $n$ i.i.d.\ samples ${\bf Z}\eqdef\{(X_i,Y_i)\}_{i=1}^n$ from $f$,  one first constructs estimators of $\Sigma_x$, $\Sigma_y$ and $\Sigma_{xy}$, denoted by $\hat\Sigma_x$, $\hat\Sigma_y$ and $\hat\Sigma_{xy}$, respectively, and then construct estimators of $A$ and $B$ (denoted by $\hat A$ and $\hat B$, respectively) from $\hat\Sigma_x$, $\hat\Sigma_y$, and $\hat\Sigma_{xy}$ in the same way as in~\eqref{cca:A_B}. 
In Section~\ref{sec:num}, we will provide some examples of constructing $\hat\Sigma_x$, $\hat\Sigma_y$, and $\hat\Sigma_{xy}$. Based on $\hat A$ and $\hat B$, one then solves (\ref{cca:R}) with the Rayleigh quotient $\textsf{R}(\cdot)$ replaced by its sample version $\textsf{R}_n(\cdot;{\bf Z})$, which is defined as
\[\textsf{R}_n(\theta;{\bf Z}) \eqdef \frac{\theta^\T \hat A \theta}{\theta^\T\hat B \theta},\quad\forall\,\theta\in\rset^{p}.\]

To guarantee that the Rayleigh quotient $\textsf{R}_n(\cdot;{\bf Z})$ is well-defined, we maintain the following assumption throughout this work.

\begin{assumption}\label{H0}
	For all $\theta\in\rset^{p}$, $|\theta^\T \hat A \theta | \leq \theta^\T\hat B \theta$.
\end{assumption}

\begin{remark}
	H\ref{H0} implies that  $\theta^\T \hat A \theta=0$ whenever $\theta^\T \hat B \theta=0$, in which case we have $\textsf{R}_n(\theta;{\bf Z})=0/0=0$. We note that H\ref{H0} naturally holds when $\hat\Sigma_x$, $\hat\Sigma_y$, and $\hat\Sigma_{xy}$ are sample covariance matrices. 
	Indeed, if $\hat\Sigma_x = n^{-1} \sum_{i=1}^n X_iX_i^\T$, $\hat\Sigma_y = n^{-1} \sum_{i=1}^n Y_iY_i^\T$, and $\hat\Sigma_{xy} = n^{-1} \sum_{i=1}^n X_iY_i^\T$, then for $\theta = (v_x^\T,v_y^\T)^\T$, and by Cauchy-Schwarz's inequality
	\begin{multline*}
		\textstyle |\theta^\T \hat A \theta| = \left| \frac{2}{n}\sum_{i=1}^n v_x^\T X_iY_i^\T v_y \right| \leq \frac{2}{n}\left\{\sum_{i=1}^n \pscal{v_x}{X_i}^2\right\}^{1/2}\left\{\sum_{i=1}^n\pscal{v_y}{Y_i}^2\right\}^{1/2}\\
		\textstyle  = 2 \sqrt{v_x^\T \hat\Sigma_x v_x} \sqrt{v_y^\T \hat\Sigma_y v_y} \leq v_x^\T \hat\Sigma_x v_x + v_y^\T \hat\Sigma_y v_y = \theta^\T \hat B \theta.
	\end{multline*}
\end{remark}

It is worth mentioning that in high-dimensional regimes where $p>n$, the constructed estimators $\hat\Sigma_x$ and $\hat\Sigma_y$ (e.g., sample covariance matrices) are usually singular, thereby making a direct maximization of $\textsf{R}_n$ challenging. Similarly, other classical CCA algorithms based on eigen-decomposition of $\hat B^{-1}\hat A$, or the singular value decomposition of $\hat\Sigma_x^{-1/2}\hat\Sigma_{x,y}\hat\Sigma_y^{-1/2}$ (see e.g., (\cite{mardia:etal:79,andrew:etal:13})) are also difficult to use under these regimes. Furthermore, these classical methods do not yield sparse estimates of the canonical correlation vectors.

(\cite{tan2018sparse}) addressed these issues by maximizing $\mathsf{R}_n(\cdot;{\bf Z})$ under a sparsity constraint. The authors show that this maximization problem can be solved provided that a good initial value that is sufficiently close to global maxima is provided. However, finding such a good initial value is very costly. Furthermore, the Rayleigh quotient typically admits several local maxima (as well as local minima and saddle points) that correspond to other canonical vectors, making direct maximization of $\textsf{R}_n$ very challenging. 

\subsection{A Quasi-Bayesian approach}
We propose a quasi-Bayesian framework that  turns maximizing the Rayleigh quotient  into a Bayesian procedure. More precisely, we propose  using 
\begin{align}
	\theta\mapsto \sigma_n\textsf{R}_n(\theta;{\bf Z})\label{bayes:quasi-L}
\end{align}
as  the quasi-log-likelihood, where $\sigma_n > 0$ is a scaling parameter. 
We combine this quasi-log-likelihood with a spike-and-slab prior distribution, which is a common choice for Bayesian sparse modeling (\cite{george:mcculloch:97}).  Specifically, given a variable selection parameter $\delta\in\Delta\eqdef\{0,1\}^{p}$, we let the conditional distribution of $\theta$ given $\delta$ be 
\begin{equation}\label{prior:1}
	\pi(\theta\vert \delta)=\prod_{j=1}^{p}\pi(\theta_j\vert \delta),\;\; \mbox{  where }\; \; \theta_j\vert \delta  \; = \;\theta_j\vert \delta_j \sim \left\{ \begin{array}{ll} \textbf{N}(0,\rho_1^{-1}), & \;\mbox{ if }\; \delta_j=1\\ \textbf{N}(0,\rho_0^{-1}), & \;\mbox{ if }\;\delta_j=0\end{array}\right.,
\end{equation}
where $\rho_0>\rho_1>0$ are precision parameters.  Given some parameter $\mathsf{u}>1$ and integer $s\geq 1$, the prior distribution of $\delta$ is taken as the independent product of Bernoulli distribution $\textsf{Ber}(1/(1+p^{\mathsf{u}}))$ conditioned to stay in the set $\Delta_s\eqdef\{\delta\in\Delta:\;\|\delta\|_0\leq s\}$. More specifically,
\begin{equation}\label{prior:2}
	\pi(\delta) \propto\textbf{1}_{\Delta_s}(\delta)\prod_{j=1}^p \left(\frac{1}{1+p^{\mathsf{u}}}\right)^{\delta_j}\left(\frac{p^{\mathsf{u}}}{1+p^{\mathsf{u}}}\right)^{1-\delta_j} \;\;\;\propto  \;\textbf{1}_{\Delta_s}(\delta) \left(\frac{1}{p^{\mathsf{u}}}\right)^{\|\delta\|_0}, \;\;\;\;\;\; \forall\,\delta\in\Delta.
\end{equation}
If we combine  the spike-and-slab  prior with the quasi-log-likelihood in~\eqref{bayes:quasi-L}, we then obtain the quasi-posterior distribution
\begin{equation}\label{post:Pi:sCCA}
	\Pi(\delta,\rmd\theta\vert {\bf Z}) \propto  \textbf{1}_{\Delta_s}(\delta) \left(\frac{1}{p^{\mathsf{u}}}\sqrt{\frac{\rho_1}{\rho_0}}\right)^{\|\delta\|_0}\exp\left(-\frac{\rho_1}{2}\|\theta_\delta\|_2^2 -\frac{\rho_0}{2}\|\theta-\theta_\delta\|_2^2+ \sigma_n \textsf{R}_n(\theta_\delta;{\bf Z})\right)\rmd \theta,
\end{equation}
where $\theta_\delta$ is the component-wise product of $\theta$ and $\delta$, $\|\cdot\|_2$ is the Euclidean norm. Note that in this posterior distribution, the parameter $\theta$ is typically dense. However, since $\delta$ is sparse, so is $\theta_\delta$. We note that the Rayleigh quotient $\textsf{R}_n$ can take value $+\infty$ when its numerator is non-zero while its denominator is zero. If this happens over a set with non-zero Lebesgue measure, then (\ref{post:Pi:sCCA}) is not well-defined. H\ref{H0} rules out these cases.

%
%
%

The  spike-and-slab prior shown in (\ref{prior:1}) and (\ref{prior:2}) is fairly standard, and goes back at least to (\cite{george:mcculloch:97}). However the way it is combined with the pseudo-likelihood to yield (\ref{post:Pi:sCCA}) is nonstandard, and  follows from (\cite{AB:19}). The key feature of this approach is that  the parameter $\theta$ enters the quasi-likelihood only through its sparsified form $\theta_\delta$ (see (\ref{post:Pi:sCCA})).   This   decouples the active components (namely those corresponding to $\delta_j=1$) and the non-active components (namely those corresponding to $\delta_j=0$), and is particularly attractive from the computational standpoint. The approach should be viewed as an approximation of the point-mass spike-and-slab prior (\cite{mitchell:beauchamp:88}), using  the pseudo-prior device in (\cite{carlin:chib:95}). We refer the reader to (\cite{AB:19}) for more details. However, we point out that the posterior contraction theory developed in (\cite{AB:19}) cannot be applied to our setting.


\subsubsection{Hyper-parameter tuning}
The posterior distribution $\Pi$ is very robust to the choice of $\rho_1$ and $\mathsf{u}$, and we recommend choosing $\rho_1\approx 1$ and $\mathsf{u} \in (1,2]$ for best performance.  The parameter  $\rho_0$  has no effect on the statistical recovery of the selected components of $\theta$, but can adversely impact the MCMC mixing if its value is too large. We suggest setting $\rho_0\sim n$,  in order to match the posterior variance of the selected components that are actually zero (false-positives), and the posterior variance of the true-negatives. 

The sparsity level $s$ is an upper-bound on the true sparsity of the signal, which is typically unknown. We observe that if $\delta_1,\ldots,\delta_p \stackrel{i.i.d.}{\sim}\textsf{Ber}(1/(1+p^{\mathsf{u}}))$, then by Chernoff's inequality (see e.g., \cite[Theorem 2.3.1]{vershynin:18}), for any $s_0\geq \exp(1)$, we have
$\PP(\|\delta\|_0 >s_0) \leq \left({1}/{p}\right)^{(\mathsf{u-1})s_0}.$
This suggests simply choosing $s = p$ in (\ref{prior:1}), and the resulting prior distribution would still automatically concentrate on sets $\Delta_{s_0}$, for $s_0$ small.  We made this choice in all our numerical implementations. We found that the resulting posterior distribution is always automatically sparse, and  learns the true sparsity of the signal. However, for the theoretical analysis of the method we will assume that a sparsity level $s$ is given such that $n\geq c_0 s\log(p)$, for some absolute constant $c_0$. We discuss the choice of $\sigma_n$ below after the statement of Theorem \ref{thm:1}.

\subsection{Connection with simulated annealing}\label{sec:sa}
Our methodology  can be viewed as a principled version of simulated annealing algorithm (\cite{kirkpatrick:etal:83,bertsimas:tsitsiklis:93}) for computing the  \textsf{Rifle} estimator of (\cite{tan2018sparse}). Given $s\geq 1$, let $\Theta_s\eqdef\{\theta\in\rset^p:\;\|\theta\|_0\leq s\}$. Let $\sigma_t>0$ be given such that $\lim_{t\to\infty} \sigma_t = +\infty$,  and define
\begin{equation}\label{ann:dist}
	\Pi_t(\rmd \theta) \propto e^{\sigma_t \textsf{R}_n(\theta;{\bf Z})} \textbf{1}_{\Theta_s}(\theta)\rmd \theta,\end{equation}
where $\rmd\theta$ denotes the extension of the Lebesgue measure to the set $\Theta_s$. The maximization problem tackled by the authors of \textsf{Rifle} in  (\cite{tan2018sparse}) is $\max_{\theta\in\Theta_s} \textsf{R}_n(\theta;{\bf Z})$. A simulated annealing solution to this problem consists in simulating a non-homogeneous Markov chain with sequence of transition kernels $\{\M_k,\;k\geq 1\}$, such that $\M_k$ has invariant distribution $\Pi_{t_k}$. As $\sigma_{t_k}\to\infty$, the distribution $\Pi_{t_k}$ puts most of its probability mass around the global modes of $\textsf{R}_n$, and the resulting Markov chain behaves similarly (under appropriate conditions).  There are however several limitations to simulated annealing  in this particular setting. First, the set $\Theta_s$ is a union of a large number of subsets with varying dimensions. Therefore, sampling from $\Pi_{t_k}$ (that is, designing a good Markov kernel $\M_k$ with invariant distribution $\Pi_{t_k}$) is actually non-trivial. Second, the convergence of simulated annealing is known to be highly dependent on the choice of the sequence $\{\sigma_{t_k},\;k\geq 1\}$. Our approach circumvents the first issue by working with a relaxation of $\Theta_s$, using the spike-and-slab prior. We circumvent the second issue by connecting the annealing schedule to the sample size $n$ ($\sigma_{t_k}=\sigma_n$, see details below), in such a way that the fluctuations in the resulting distribution $\Pi_{t_k}$ matches the statistical uncertainty of the underlying CCA problem. 


\subsection{Rate of convergence}\label{sec:post:contr}
Although the Rayleigh quotient $\textsf{R}_n(\cdot;{\bf Z})$ may possess multiple  local modes, we show in this section that most of the probability mass of the quasi-posterior distribution $\Pi(\cdot\vert {\bf Z})$ are located around $\{\pm\theta_\star\}$. For $M,N\in\rset^{q\times q}$, 
we define
\[\pscal{M}{N}_{\mathsf{F}} \eqdef \mathsf{Tr}(M^\T N),\;\;\;\|M\|_{\mathsf{F}} \eqdef\sqrt{\pscal{M}{M}_{\mathsf{F}}},\;\;\mbox{ and } \;\;\|M\|_{\textsf{op}} \eqdef \sup_{u\in\rset^q:\;\|u\|_2=1}\; \|M u\|_2.\]
For $J\subseteq[1:q]\eqdef\{1,\ldots,q\}$, let $M_{J,J}$ denote the submatrix $(M_{ij})_{i,j\in J}$. Given  $k\geq 1$,  we let
\[\lambda_{\mathsf{min}}(M,k)\;\eqdef \;\min_{u\in\rset^q:\;\|u\|_2=1,\|u\|_0\leq k} \;u^\T M u,\;\;\mbox{ and }\;\;  \lambda_{\mathsf{max}}(M,k)\;\eqdef\; \max_{u\in\rset^q:\;\|u\|_2=1,\|u\|_0\leq k} \;u^\T M u.\]
Given an integer $\alpha\geq 1$, we set
\[\lambda_{\mathsf{max}}^{(\alpha)}(M,s)\;\eqdef \max_{J\subseteq[1:q]:\;\|J\|_0 = s}\;\;\max_{\stackrel{A\in\rset^{s\times s}:\;\|A\|_{\mathsf{F}}=1}{ \;\mathsf{Rank}(A)\leq \alpha}}\;\;\; \left|\pscal{M_{J,J}}{A}\right|.\]
We first make the following basic assumption without which the sparse CCA problem would not be well defined.

\begin{assumption}\label{H1}
	The joint density $f$ possesses positive definite covariance matrices $\Sigma_x$, $\Sigma_y$, and $\Sigma$, and a principal  canonical vector pair $\theta_\star=(v^\T_{x\star},v^\T_{y\star})^\T$, ($\|\theta_\star\|_2=1$) with density level\footnote{Throughout this work, the density level of a vector refers to the proportion of its non-zero elements.}  $s_\star \eqdef \|\theta_\star\|_0$. Furthermore, the  difference between the largest and second largest eigenvalue of $S \eqdef B^{-1/2} \Sigma B^{-1/2}$ (denoted by $\textsf{gap}$), is positive.
\end{assumption}

\medskip

Our main assumption  on the data generation process is the following.

\begin{assumption}\label{H2}
	The dataset ${\bf Z}\eqdef\{(X_i,Y_i)\}_{i=1}^n$ and the  integer $s\geq s_\star$ are such that  the estimators  $\hat \Sigma_x$, $\hat \Sigma_y$, $\hat \Sigma$ satisfy the following.
	\begin{enumerate}
		\item For some absolute constants $0<\underline{\kappa}\leq \bar \kappa$, 
		\[\min\left(\lambda_{\textsf{min}}(\hat \Sigma_x,s+s_\star),\; \lambda_{\textsf{min}}(\hat \Sigma_y,s+s_\star),\; \lambda_{\textsf{min}}(\hat \Sigma,s+s_\star)\right)  \geq \underline{\kappa},\]
		\[\max\left(\lambda_{\textsf{max}}(\hat \Sigma_x,s+s_\star),\; \lambda_{\textsf{max}}(\hat \Sigma_y,s+s_\star), \lambda_{\textsf{max}}(\hat \Sigma,s+s_\star) \right)  \leq \bar{\kappa}.\]
		\item For some constant $\r_1$ (depending possibly on $n,p$), 
		\[\max\left(\lambda^{(2)}_{\textsf{max}}(\hat \Sigma_x- \Sigma_x,s+s_\star),\; \lambda^{(2)}_{\textsf{max}}(\hat \Sigma_y - \Sigma_y,s+s_\star),\; \lambda^{(2)}_{\textsf{max}}(\hat \Sigma- \Sigma,s+s_\star)\right)  \leq \r_1.\]
	\end{enumerate}
\end{assumption}


\begin{theorem}\label{thm:1}
	Assume H\ref{H0}-H\ref{H2}, and suppose that $p\geq \max(c_0, s_\star\exp(1))$, for some absolute constant $c_0$. Choose $\sigma_n$ such that $1\leq \sigma_n\leq p$, and $\mathsf{u}>1$ such that $p^{\mathsf{u}-1}>2$.  Set 
	\begin{equation}\label{cv:rate}
		\epsilon\eqdef \frac{\r_1}{\mathsf{gap}}.
	\end{equation}
	There exists some absolute constant $C_0$  that depends only on  $\underline{\kappa}$ and $\bar\kappa$, such that the following holds. For all  $M>C_0$ such that 
	\begin{equation}\label{cond:ss}
		\frac{M^2}{8 \textsf{gap}}\left(\frac{\underline{\kappa}}{\bar\kappa}\right)^2\sigma_n\r_1^2 \geq s_\star(\mathsf{u}+1)\log(p),\end{equation}
	we have
	\begin{equation*} \Pi\left((\delta,\theta):\;\left\| \frac{\theta_\delta\theta_\delta^\T}{\|\theta_\delta\|_2^2} - \theta_\star\theta_\star^\T \right\|_{\mathsf{F}} > 	M \epsilon \vert {\bf Z}\right) \; \leq 2e^{-\frac{M^2}{8 \textsf{gap}}\left(\frac{\underline{\kappa}}{\bar\kappa}\right)^2 \sigma_n \r_1^2} \leq \frac{2}{p^{s_\star(\mathsf{u}+1)}}.
	\end{equation*}
\end{theorem}
\begin{proof}
	See Section S-1.1 in supplementary material.
\end{proof}

The main conclusion of the theorem is that the posterior $\Pi(\cdot\vert {\bf Z})$ contracts around $\theta_\star\theta_\star^\T$ at the rate at least $M\epsilon$. 
Furthermore, setting
\[\widehat{\mathcal{P}} \eqdef \int_{\Delta_s\times\rset^p} \frac{\theta_\delta\theta_\delta^\T}{\|\theta_\delta\|_2^2} \Pi(\rmd\delta,\rmd\theta\vert{\bf Z}),
\]
the result implies that $\widehat{\mathcal{P}} $ (as a frequentist estimator) estimates  $\theta_\star\theta_\star^\T$ at the rate $M\epsilon$. Indeed, we have
\begin{equation}\label{eq:F}
	\left\| \widehat{\mathcal{P}} - \theta_\star \theta_\star^\T\right\|_{\mathsf{F}}  \leq  \int_{\Delta_s\times \rset^p} \left\| \frac{\theta_\delta\theta_\delta^\T}{\|\theta_\delta\|_2^2} - \theta_\star\theta_\star^\T\right\|_{\mathsf{F}} \Pi\left(\rmd\delta,\rmd\theta\vert {\bf Z}\right) \leq M\epsilon + 4e^{-\frac{M^2}{8 \textsf{gap}}\left(\frac{\underline{\kappa}}{\bar\kappa}\right)^2 n \r_1^2}.\end{equation}

We note that Theorem \ref{thm:1} applies to any  given dataset ${\bf Z}$ and estimators $\hat \Sigma_x$, $\hat \Sigma_y$ that satisfy H\ref{H2}, regardless of how they are formed. In the particular case where $\hat\Sigma_x, \hat\Sigma_y$ and $\hat\Sigma$ are covariances matrices, we show in Proposition \ref{prop:H2}  below that if $f$ is a sub-Gaussian distribution, then H\ref{H2} holds with high probability. Furthermore $\r_1 = C_0\sqrt{(s+s_\star)\log(p)/n}$. In that case the condition in (\ref{cond:ss}) becomes
\[\frac{M^2}{8 \textsf{gap}}\left(\frac{\underline{\kappa}}{\bar\kappa}\right)^2 C_0^2 \left(\frac{\sigma_n}{n}\right)(s+s_\star)\log(p) \geq s_\star(\mathsf{u}+1)\log(p),\]
which is easily satisfied when the scaling parameter $\sigma_n$ satisfies $n=O(\sigma_n)$, as $n\to\infty$.  In this case the convergence rate of $\widehat{\mathcal{P}}$  towards $\theta_\star\theta_\star^\T$ is 
\begin{equation}\label{rate:scov}
	\epsilon = \frac{1}{\mathsf{gap}} \sqrt{\frac{(s+s_\star) \log(p)}{n}},\end{equation}
which achieves the minimax rate of the CCA problem, as derived in (\cite{gao2015}), by taking $s$ as some constant multiple of $s_\star$. Further increasing $\sigma_n$ has no impact on this rate, but of course, makes $\Pi$ more concentrated around the modes of $\textsf{R}_n(\cdot;{\bf Z})$, thereby making the MCMC computation more challenging. This suggests that the choice $\sigma_n\propto n$ is the right scaling.

\begin{rem}
	The discussion so far has focused on estimating the projector $\theta_\star\theta_\star^\T$. If the vector $\theta_\star$ itself is needed, we are able to construct an estimator of $\theta_\star$ from the projector estimator $\widehat{\mathcal{P}}$. Specifically, let $v_1(\widehat{\mathcal{P}})$ denote the leading eigenvector of $\widehat{\mathcal{P}}$, then from the Davis-Kahan theorem (see e.g., \cite[Theorem 4.5.5]{vershynin:18}), we have
	\begin{equation}
		\min\left( \|v_1(\widehat{\mathcal{P}})- \theta_\star\|_2, \;\|v_1(\widehat{\mathcal{P}}) + \theta_\star\|_2 \right)\le 2^{3/2} \| \widehat{\mathcal{P}} - \theta_\star \theta_\star^\T \| \le 2^{3/2} \| \widehat{\mathcal{P}} - \theta_\star \theta_\star^\T \|_{\mathsf{F}},
	\end{equation}
	and $\|\widehat{\mathcal{P}} - \theta_\star \theta_\star^\T \|_{\mathsf{F}}$ can be bounded as in \eqref{eq:F}.
\end{rem}

\subsubsection{On Assumption H\ref{H2}}
It is well-known that Assumption H\ref{H2}-(1) holds true in the particular case of covariance matrices of sub-Gaussian random vectors, provided that the sample size satisfies $n\geq c_0 (s+s_\star)\log(p)$, for some absolute constant $c_0$. See for instance \cite{raskutti:etal:10}~Theorem 1, or \cite{gao:etal:17}~Lemma 6.5 for the Gaussian case, and \cite{rudelson:zhou:13}~Theorem 3.2 for more general sub-Gaussian distributions. Under roughly the same sample size conditions, H\ref{H2}-(2) is also known to hold as we show below.

\begin{proposition}\label{prop:H2}
	Suppose that ${\bf Z}\eqdef\{(X_i,Y_i)\}_{i=1}^n$ are i.i.d. random vectors from a mean-zero sub-Gaussian distribution $f$, with sub-Gaussian norm $K\eqdef \sup\{ \|\pscal{Z}{u}\|_{\psi_2},\; u\in\rset^p,\;\|u\|_2=1\}$, where $\|\cdot\|_{\psi_2}$ refers to the sub-Gaussian norm of a random variable. Let $\hat\Sigma_x=n^{-1}\sum_{i=1}^n X_iX_i^\T$, $\hat\Sigma_y=n^{-1}\sum_{i=1}^n Y_iY_i^\T$, and $\hat\Sigma=n^{-1}\sum_{i=1}^n Z_iZ_i^\T$. There exist absolute constants $c_0,C>1$, such that for all $1\leq s\leq p$, and all $n\geq 4c_0 s\log(p)$,
	\[\max\left(\lambda^{(\alpha)}_{\textsf{max}}(\hat \Sigma_x- \Sigma_x,s),\; \lambda^{(\alpha)}_{\textsf{max}}(\hat \Sigma_y - \Sigma_y,s),\; \lambda^{(\alpha)}_{\textsf{max}}(\hat \Sigma- \Sigma,s)\right)  \leq C K^2 \lambda_{\textsf{max}}(\Sigma,s) \sqrt{\frac{c_0\alpha s\log(p)}{n}},\]
	with probability $1 - 2p^{-(c_0-1)s}$. 
\end{proposition}
\begin{proof}
	See Section S-2 in supplementary material.
\end{proof}

\subsubsection{Bayesian inference}\label{sec:BI}
We have developed a method that employs a quasi-posterior distribution to produce a frequentist estimator. The idea of using a Bayesian framework to produce frequentist estimators is of course well-established in statistical decision theory (\cite{robert:BC}).  The extension to quasi-likelihood functions is also not new  (\cite{mcallester:99,catoni:01,chernozhukov:hong03}). An important statistical question here is whether one can use the full quasi-posterior distribution $\Pi(\cdot \vert{\bf Z})$ to carry inference on $\theta_\star\theta_\star^T$, for instance through credible sets. The difficulty is the lack of calibration  of the quasi-likelihood function (we could easily replace  $\sigma_n$ by $2\sigma_n$ as a scaling factor in the Rayleigh quotient). To address this issue some authors have developed post-processing methods to match samples from the quasi-posterior distribution to the corresponding frequentist central limit theorem distribution (\cite{bissiri16, benjamin14}). However these methods rely crucially on the Bernstein-von Mises theorem and the central limit theorem that are only well-understood  in fixed-dimensional settings.  Extending these ideas  to  the (high/growing)-dimensional setting remains largely open. We leave this question as a possible future research. Currently we do not advocate the use of our quasi-posterior distribution for Bayesian inference on $\theta_\star$.

\section{Computation using Markov Chain Monte Carlo}\label{sec:mcmc}
As shown in Section~\ref{sec:post:contr}, by re-scaling (annealing) the Rayleigh quotient function, we have created a posterior distribution $\Pi(\cdot\vert {\bf Z})$ that puts most of its probability mass around its global mode (located near $\{\pm\theta_\star\}$). However, the annealing also significant decreases the accessibility of the global mode starting from other parts of the space. To effectively deal with this configuration, we propose a Markov Chain Monte Carlo sampling strategy based on simulated tempering~(\cite{geyer:95,liu2008monte}). 
Given $K$ temperatures $1=t_1<t_2<\ldots<t_K$, and $K$ positive weights $c_1,\ldots,c_K$, we introduce an extended distribution on $ \mathsf{X} \eqdef \Delta\times \rset^p\times \{1,\ldots,K\}$, which is
\begin{equation}\label{post:Pi:sCCA:temp}
	\bar\Pi(\delta,\rmd\theta,k \vert {\bf Z}) \propto \frac{1}{c_k}\exp\left(\frac{\mathsf{a}}{t_k} \|\delta\|_0-\frac{\rho_1}{2t_k}\|\theta_\delta\|_2^2 -\frac{\rho_0}{2t_k}\|\theta-\theta_\delta\|_2^2+\frac{\sigma_n}{t_k} \textsf{R}_n(\theta_\delta;{\bf Z})\right)\rmd \theta.
\end{equation}
We recover the distribution (\ref{post:Pi:sCCA}) as the conditional distribution of $(\delta,\theta)$ given $k=1$ in (\ref{post:Pi:sCCA:temp}). To sample from  (\ref{post:Pi:sCCA:temp}), we use a simulated tempering Metropolis-Hastings-within-Gibbs strategy that is described in Algorithm 1 in the supplementary material S-3.  The algorithm is very fast and scales well with the dimension $p$, and  iteration $k$ of the algorithm  has computational cost  $O(p\|\delta^{(k)}\|_0^2)$.

Algorithm 1 generates a Markov chain $\{X^{(t)},\;t\geq 0\}$, where $X^{(t)} = (\delta^{(t)},\theta^{(t)},k^{(t)}) \in \mathsf{X}$ that is phi-irreducible aperiodic with invariant distribution given by (\ref{post:Pi:sCCA:temp}). The pairs $(\delta^{(t)},\theta^{(t)})$ at times $t$ where $\{k^{(t)}=1\}$ then give the desired approximate samples from $\Pi(\cdot\vert {\bf Z})$.
For the investigation of mixing of the algorithm, please see the supplementary material.

\section{Numerical studies}\label{sec:num}
We perform a simulation study that compares our approach to the frequentist methods \textsf{Rifle} in \cite{tan2018sparse} and \textsf{mixedCCA} in \cite{yoon:etal:18}. We investigate the behavior of these methods in two settings: (i) continuous datasets, where we use sample covariance matrix estimator and (ii) mixed datasets, where we use Kendall's-tau-based estimator as proposed in \cite{yoon:etal:18}. The \textsf{Python} codes for our method is available from \url{https://github.com/rachelwho/Sparse-CCA}.

\subsection{Simulated data generation}\label{sec:data:model}
We simulate  the datasets using the following model  from \cite{tan2018sparse}. Specifically, we let $p_x = p_y = p/2$, and  consider two $(p/2)$-dimensional random vectors $X$ and  $Y$ with joint distribution $(X, Y) \sim \textbf{N}(0, \Sigma)$. Here we let 
\begin{align*}
	\Sigma  = 
	\begin{pmatrix}
		\Sigma_{x} & \Sigma_{xy}\\
		\Sigma_{xy}^\T & \Sigma_{y}
	\end{pmatrix}\quad \mbox{and}
	\quad \Sigma_{xy} = \frac{\lambda_1  \Sigma_{x} v_{x\star} v_{y\star}^\T \Sigma_{y}}{\sqrt{v_{x\star}^\T  \Sigma_{x} v_{x\star}} \sqrt{v_{y\star}^\T \Sigma_{y} v_{y\star}} } ,
\end{align*}
where $0< \lambda_1 <1$ is the largest generalized eigenvalue, and $v_{x\star}$ and $v_{y\star}$ are the principal canonical vectors. The structures of $\Sigma_x$ and $\Sigma_y$ vary across different experimental settings, and will be described in the subsequent sections. 
Clearly, $(v_{x\star},v_{y\star})$ is the maximizer of the Rayleigh quotient in~\eqref{cca:R}, and $\lambda_1$ is the maximum value.  Then we generate  $n$ samples $\mathbf{Z}=\{(X^{(i)}, Y^{(i)})\}_{i = 1}^n$ 
from $\textbf{N}(0, \Sigma)$. 

\subsection{Comparison with other methods}
We compare our method to two other methods, namely \textsf{Rifle} \cite{tan2018sparse} and \textsf{mixedCCA} \cite{yoon:etal:18}. We investigate the behaviors of these methods in two settings. In the first setting, we use continuous datasets and compare our method with both \textsf{Rifle} and \textsf{mixedCCA}. In the second setting, we use mixed datasets and compare our method with \textsf{mixedCCA} (since \textsf{Rifle} is only designed for the continuous datasets). 

\subsubsection{Description of \textsf{Rifle} and \textsf{mixedCCA}}

Before presenting our experimental results, let us briefly describe the other two methods, namely \textsf{Rifle} and \textsf{mixedCCA}.  \textsf{Rifle} is a two-stage algorithm, where in the first stage, it (approximately) solves a  convex relaxation of the problem in (\ref{cca:probl:0}) to produce an initial estimate of the singular vectors $(v_x^\T,v_y^\T)^\T$, which  
are then refined in the second stage using gradient ascent on the Rayleigh quotient $\textsf{R}_n(\cdot;\mathbf{Z})$, with a truncation step such that only the $m$ entries with the largest absolute values are kept  (and  the remaining entries are set to zero). Here $m$ is a user-specified parameter that indicates the desired sparsity level of the estimated principle canonical vectors $(v_x,v_y)$ --  similar to $s$ above.  Note that since the first stage involves solving a matrix optimization problem, its computational time is typically much higher than that of the second stage. 
As a different approach, \textsf{mixedCCA} proposes a novel and robust estimator $\hat\Sigma$ for the covariance matrix $\Sigma$, namely the Kendall's-tau-based estimator, and estimates the canonical vectors $(v_x,v_y)$ by solving the following convex problem: 
\begin{align}\label{est:yoon}
	\max_{v_x,v_y} v_x^\T \hat\Sigma_{xy} v_y - \lambda_1 \|v_x \|_1 - \lambda_2 \|v_y\|_1,\;\;\mbox{ s.t.}\;\;v_x^\T \hat\Sigma_{x} v_x \leq 1,\;\;v_y^\T \hat\Sigma_{y} v_y \leq 1,
\end{align}
where $\lambda_1$ and $\lambda_2$ are positive regularization parameters that need to be selected. Problem (\ref{est:yoon}) is then solved via a sequence of LASSO problems.

\subsubsection{Comparison with continuous datasets}\label{sec:cont}


We randomly generated 100 continuous datasets using the model in Section~\ref{sec:data:model}, with the covariance matrices $\Sigma_x$ and $\Sigma_y$ constructed in a similar way to \cite{yoon:etal:18}. Specifically, we set the sample size $n=200$ and the dimension $p = 500$, and let $\Sigma_{x}$ and $\Sigma_{y}$ have the same structure, namely a block-diagonal matrix with five blocks of dimensions $\{d_1, ..., d_5\}$, respectively, and the $(j,j')$-th element of each block takes value $0.7^{|j-j'|}$. We set $\{d_1, ..., d_5\} = \{25, 50, 83, 50, 42\}$ for $\Sigma_{x}$ and $\{d_1, ..., d_5\}=\{83, 50, 62, 31, 24\}$ for $\Sigma_{y}$. In addition, we let $\lambda_1=0.8, (v_{x\star})_j =(v_{y\star})_j= 1/\sqrt{3}$ for $j \in \{ 1, 6, 11\}$, and $(v_{x\star})_j = (v_{y\star})_j = 0$ otherwise. Therefore, the true density level  is $s_\star = 6$. In constructing the Rayleigh quotient $\textsf{R}_n(\cdot;\mathbf{Z})$, we used the sample covariance matrices as estimators of $\Sigma_x$, $\Sigma_y$ and $\Sigma_{xy}$.


In Algorithm~2, 
we let the set of temperatures be $ \{1, 1/0.9,  1/0.8, 1/0.7, 1/0.6\}$, and only recorded the  iterations corresponding to temperature 1. For comparison, we used the implementation of \textsf{Rifle} in the $\mathsf{R}$ package $\mathsf{Rifle}$, and set the parameter $m=2s_\star = 12$. 
As pointed out in~\cite{tan2018sparse}, the first stage is computationally expensive to run. In addition,  we empirically found 
that  when the sample size $n$ is not sufficiently large, either the estimated $v_x$ or $v_y$ from the first stage of \textsf{Rifle} is often zero vector, which caused us serious problems in running the second stage. 
Because of these issues, we evaluated separately the two stages of  $\mathsf{Rifle}$, which we call  $\mathsf{Rifle}1$ and $\mathsf{Rifle}2$, respectively.  We ran $\mathsf{Rifle}1$ 
with default parameters,  and ran $\mathsf{Rifle}2$ 
starting from a solution generated by perturbing the ground-truth $(v_{x\star}^\T, v_{y\star}^\T)^\T$, where the perturbation was drawn from a  centered Gaussian with standard deviation $0.2$. 
We used the implementation of $\textsf{mixedcca}$ in  the \textsf{R} package $\textsf{mixedCCA}$, where $\lambda_1$ and $\lambda_2$ were selected using two different criteria, namely BIC1 and BIC2. 
For this reason, we shall call the resulting algorithms $\textsf{mixedCCA-BIC1}$ and $\textsf{mixedCCA-BIC2}$, respectively. 
All the other parameters in $\textsf{Rifle}$ and $\textsf{mixedCCA}$ were set to the default values in the \textsf{R} packages. Both our algorithm and $\textsf{mixedCCA}$ used the starting point found in the \textsf{R} package of $\textsf{mixedCCA}$. The output of each algorithm was normalized to have unit Euclidean norms. 

\paragraph{Comparison of running times.} We first compare the computational efficiency of different algorithms. Since these algorithms converge to possibly different estimators, 
we first ran each algorithm for a maximum iteration of $2000$ to obtain the ``limit point'' of the sequence generated by each algorithm,  denoted by $(\hat{v}_x^\T,\hat{v}_y^\T)^\T$. 
Then, we terminated each algorithm if it either reached $1000$ iterations  or  the estimate $({v}_x^\T,{v}_y^\T)^\T$ satisfies $\max\{|\textsf{error}(v_x)-\textsf{error}(\hat{v}_x)|,|\textsf{error}(v_y)-\textsf{error}(\hat{v}_y)|\}\le 1\times 10^{-4}$. 
As mentioned above, we treated the two stages of \textsf{Rifle} separately. We estimated the computation time for  \textsf{Rifle1}  using the default stopping criterion as in~\cite{tan2018sparse}, and estimated the running time of \textsf{Rifle2}  (starting from the perturbed ground-truth) using the termination criterion described above. 

We repeatedly ran these algorithms on 100 different simulated datasets, 
and show the averaged estimated computation times of the algorithms in Table~\ref{tab:cost}.   The results confirm the high computational cost of \textsf{Rifle}. The results also show that our proposed estimator remains computationally competitive compared to \textsf{mixedCCA}, even though it is based on MCMC.

\begin{table}[]
	\caption{The computation times of all algorithms averaged across 100 continuous datasets.}
	\centering
	\resizebox{\columnwidth}{!}{
		\begin{tabular}{|l|c|c|c|c|c|}
			\hline
			Algorithm              & {\sf Simulated tempering} & {\sf MixedCCA-BIC1} & {\sf MixedCCA-BIC2} & {\sf Rifle1} & {\sf Rifle2}\\ \hline
			Running times (s) & 12                  & 9.6          & 10.9           & 276                 & 2.2                                       \\ \hline
	\end{tabular}}
	\label{tab:cost}
\end{table}

\paragraph{Comparison of statistical efficiency.} 
We measure the quality of the estimated principle canonical vectors $v_{x}$ and $v_{y}$ using three metrics. The first one is the squared-$l_2$ errors of $v_{x}$ and $v_{y}$ to the ground-truth $v_{x\star}$ and $v_{y\star}$, respectively. Specifically, we have  
\begin{equation}\label{eq:error}
	\textsf{error}(v_x) \eqdef  \min\left( \left\|v_x- v_{x\star}\right\|_2^2, \;\left\|v_x + v_{x\star}\right\|_2^2 \right),
\end{equation}
and $\textsf{error}(v_y)$ is defined similarly.
The other two metrics are true-positive rate (\textsf{TPR}) and true-negative rate (\textsf{TNR}), which measure the quality of variable selection by the estimated $v_{x}$ and $v_{y}$. 
For $v_x$, its \textsf{TPR} and \textsf{TNR} are defined as
\begin{align}\label{eq:tpr}
	\hspace{-2ex}\textsf{TPR}(v_x) \eqdef \frac{| \{ j: (v_x)_j \neq 0, (v_{x\star})_j \neq 0 \}| }{|\{ j: (v_{x\star})_j \neq 0 \}|} \;\; \mbox{and}\;\; \textsf{TNR}(v_x) \eqdef \frac{| \{ j: (v_x)_j = 0, (v_{x\star})_j = 0 \}| }{|\{ j: (v_{x\star})_j = 0 \}|},
\end{align}
respectively, and for $v_y$, its $\textsf{TPR}$ and $\textsf{TNR}$ are defined similarly. 

\begin{figure}
	\centering
	\begin{subfigure}[b]{\textwidth}
		\centering
		\includegraphics[width=0.45\textwidth]{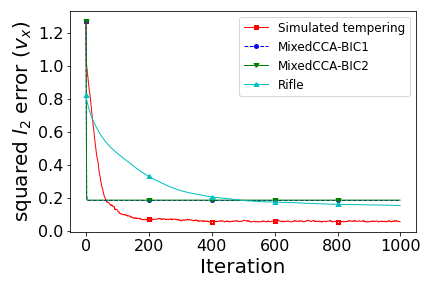}
		\includegraphics[width=0.45\textwidth]{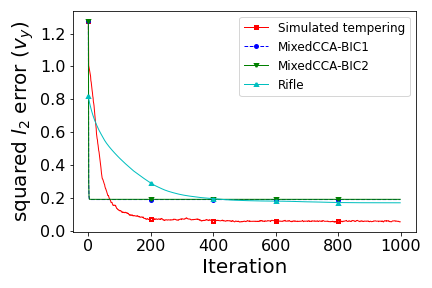}
		\vspace{-6mm}
		\caption{Squared-$l_2$ error (\textsf{error})}
		\label{fig_error}
	\end{subfigure}
	\vfill
	\begin{subfigure}[b]{\textwidth}
		\centering
		\includegraphics[width=0.45\textwidth]{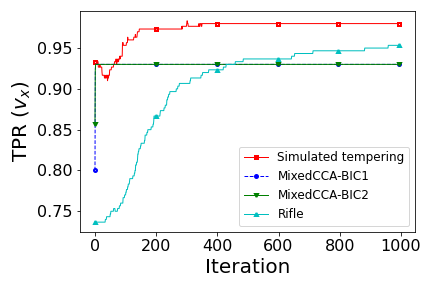}
		\includegraphics[width=0.45\textwidth]{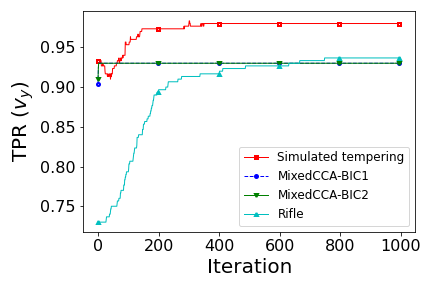}
		\vspace{-6mm}
		\caption{\textsf{TPR}}
		\label{fig_tpr}
	\end{subfigure}
	\vfill
	\begin{subfigure}[b]{\textwidth}
		\centering
		\includegraphics[width=0.45\textwidth]{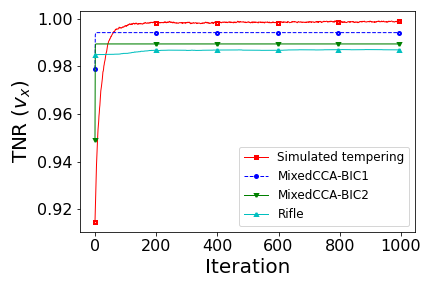}
		\includegraphics[width=0.45\textwidth]{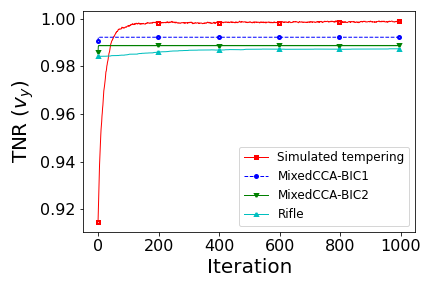}
		\vspace{-6mm}
		\caption{\textsf{TNR}}
		\label{fig_tnr}
	\end{subfigure}
	\vspace{-12mm}
	\caption{ Comparison of the quality of estimated $v_x$ and $v_y$ by all the algorithms in terms of  (a) squared-$l_2$ error (\textsf{error}), (b) \textsf{TPR} and (c) \textsf{TNR}. The results are averaged over 100 continuous datasets. To better compare \textsf{TPR} and \textsf{TNR}, we show the results starting from the first iteration, since the initial points are usually not sparse. The performances of \textsf{mixedCCA-BIC1} and \textsf{mixedCCA-BIC2} are indistinguishable on plots (a) and (b).}
	\label{fig:continuous}
\end{figure}

To estimate these metrics we run all the algorithms for 1000 iterations, well beyond their convergence times. For each algorithm, we plot the quality of the estimated $v_x$ and $v_y$ (measured by \textsf{error}, \textsf{TPR} and \textsf{TNR}) 
averaged across 100 datasets, and the results are shown  in Figure~\ref{fig:continuous}. Note that for \textsf{Rifle}, we only plot its second stage, which has a better starting point (namely, the randomly perturbed ground-truth) as compared to the other two algorithms.

From both Figure~\ref{fig:continuous} and Table~\ref{tab:cost}, we see that our algorithm  not only outperforms \textsf{Rifle}  in terms of the quality of estimated $v_x$ and $v_y$ (across all the three metrics), but also enjoys much shorter running time. 
Compared with \textsf{MixedCCA}, although our algorithm  has slightly longer computational time, the quality of estimated $v_x$ and $v_y$ from our algorithm is better, and the advantage is especially significant  in terms of \textsf{error} and \textsf{TPR}.

\subsubsection{Comparison in a mixed data setting}\label{sec:mix:data}
In many applications, particularly bio-medical ones, researchers often face the challenge that one of the variables $X$ or $Y$ is not observed directly, but only through its truncated or quantized version. Specifically,  we consider the truncated latent Gaussian copula model of  (\cite{yoon:etal:18}), which extends both the Gaussian copula model (\cite{liu:etal:09}) and the latent Gaussian copula model (\cite{fan:etal:17}). 
\begin{definition}[Gaussian copula model]
	A random vector  $Z = (Z_1, \ldots,  Z_p)^\T$ is a realization of the Gaussian copula model, if there exists a transformation $h: \rset^p \to \rset^p$ such that $h(Z) = ( h_1(Z_1),  \ldots, h_p(Z_p) )^\T \sim \textbf N (0, \Sigma)$ and for each $j = 1, \ldots, p$, transformation $h_j : \rset \to \rset$ is monotonically increasing. We write this as $Z \sim \textbf{NPN}(0, \Sigma, h)$.
\end{definition}

\begin{definition}[Truncated Gaussian copula model]
	The random vector $ (X^\T, Y^\T)^\T$, where $X \in \rset^{p_x}$ and $Y \in \rset^{p_y}$, is a realization of a latent Gaussian copula model with truncation if there exists a random vector 
	$U \in \rset^{p_y}$ such that  $(X, U) \sim \textbf{NPN}(0, \Sigma, h)$ and $Y_j = I(U_j > C_j)(U_j-C_j) + C_j $ for all $j = 1, \ldots, p_y$, where $C = (C_1, \ldots, C_{p_y})$ is a truncation parameter.  We write $(X,Y) \sim \textbf{TNPN}(0, \Sigma, h, C)$.
\end{definition}
Taking $h$ as the identity map, suppose that we are interested in the sparse CCA of $(X,U)\sim\textbf{N}(0, \Sigma)$, but we observe only independent copies of $(X,Y)$, where $Y_j= I(U_j > C_j)(U_j-C_j) + C_j$, for truncation levels $C = (C_1, \ldots, C_{p_y})$. Clearly, the classical Pearson sample covariance estimator cannot be used to estimate $\Sigma$.  Nevertheless, building on (\cite{fan:etal:17}), (\cite{yoon:etal:18}) showed that consistent estimators for $\Sigma_x$, $\Sigma_y$ and $\Sigma_{xy}$ can be constructed from independent replications of $(X,Y)$ using a Kendall's-tau covariance. Based on those estimates one can readily apply our Rayleigh quotient approach to obtain the sparse canonical correlation vectors of $\Sigma$. We compare our estimator with \textsf{MixedCCA}. In this  mixed data setting, and unlike the continuous data setting, we found out that the two methods have comparable performances, with a slight advantage to our method in terms of statistical recovery, and a slight advantage to \textsf{MixedCCA} in terms of computational speed. We illustrate this below in a low sample size regime.

We randomly generated 100  mixed datasets in a similar way as in Section~\ref{sec:cont}, except with an additional truncation step on the random vector $Y$. Specifically, we set the sample size $n=180$ and the dimension $p = 200$, and let $\Sigma_{x}$ and $\Sigma_{y}$ each have  five diagonal blocks of dimensions $\{d_1, ..., d_5\}$, respectively, and the $(j,j')$-th element of each block takes value $0.7^{|j-j'|}$. We set $\{d_1, ..., d_5\} = \{10, 20, 33, 20, 17\}$ for $\Sigma_{x}$ and $\{d_1, ..., d_5\}=\{33, 20, 25, 12, 10\}$ for $\Sigma_{y}$. In addition, we let $v_{x\star}$ and $v_{y\star}$ have the same structures as in Section~\ref{sec:cont} (so that the true density level  $s_\star=6$), and set $\lambda_1=0.8$. Let $\mathsf{truc}(\cdot;C)$ be the (elementwise) truncation operator at level $C$, such that given any vector $y$, $\mathsf{truc}(y;C)_j = y_j$ if $y_j>C$ and  $\mathsf{truc}(y;C)_j = C$ otherwise. (In particular, we can recover the continuous data setting for $C$ negatively large.) For each dataset, we generated  $n$  samples from $(X,\mathsf{truc}(U;C))$, where $(X,U)\sim\mathbf{N}(0,\Sigma)$. 

We ran Algorithm~2 with the set of temperatures $ \{1, 1/0.9,1/0.8, 1/0.7, 1/0.6\}$, that we  compare with both $\textsf{mixedCCA-BIC1}$ and $\textsf{mixedCCA-BIC2}$ in terms of  the running time and the statistical performances, as  measured  in  Section~\ref{sec:cont}. To evaluate the convergence times, we first run both algorithms for $N=10,000$  iterations to obtain their respective``limit points''. 

\begin{figure}
	\centering
	\includegraphics[width=0.45\textwidth]{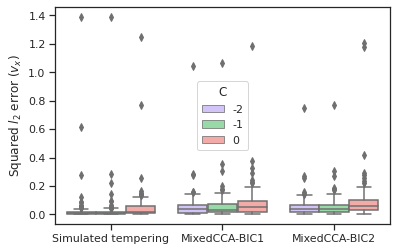}
	\includegraphics[width=0.45\textwidth]{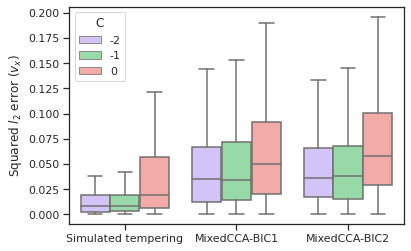}
	\vspace{-1mm}
	
	
	\caption{\label{fig:merrorx} Squared-$l_2$ error (\textsf{error}) of estimated $v_x$ by all the algorithms for different  truncation levels $C$, with outliers (Left) and without outliers (Right)}
	
\end{figure}

\begin{figure}
	\centering
	\includegraphics[width=0.45\textwidth]{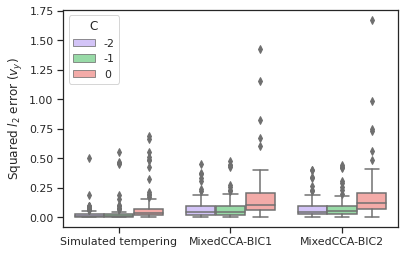}
	\includegraphics[width=0.45\textwidth]{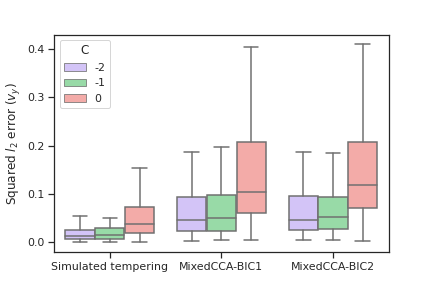}
	\vspace{-1mm}
	
	
	\caption{\label{fig:merrory} Squared-$l_2$ error (\textsf{error}) of estimated $v_y$ by all the algorithms for different  truncation levels $C$, with outliers (Left) and without outliers (Right)}
	
\end{figure}

The statistical performances of these algorithms (as measured by \textsf{error}, \textsf{TPR} and \textsf{TNR}) over the 100 mixed datasets generated as above are shown in Figure~\ref{fig:merrorx}, and Figure~\ref{fig:merrory} and Table~\ref{tab:2}.  Because \textsf{TPR} and \textsf{TNR} are discrete values, we show the results of \textsf{TPR} and \textsf{TNR} in terms of mean and standard deviation. The computation times are recorded in Table \ref{tab:mix:cost}.  
Due to the low sample size, both methods are prone to producing poor estimates that we consider as outliers. The boxplots in Figure~\ref{fig:merrorx} and Figure~\ref{fig:merrory} report the distributions of $\textsf{error}(v_x)$ and $\textsf{error}(v_y)$ respectively, with and without these outliers (by removing the points outside of the whiskers of the boxplots). 

In the low-truncation regime ($C=-2$) we recover the same conclusion as in the continuous data setting that our method outperforms \textsf{mixedCCA}. In the high-truncation setting ($C=0$), our method still slightly outperforms \textsf{mixedCCA}, particularly in  the recovery of $v_y$. The performance in terms of \textsf{TPR} and \textsf{TNR} are mostly similar, but again with a slight advantage to our method. However, here the computational time of our estimator is noticeably higher than $\textsf{mixedCCA}$ as shown in Table \ref{tab:mix:cost}. 


%
%
%

\newcommand{\wra}{.8}
\setlength{\tabcolsep}{.3em}
\begin{table}[t!]
	\centering

	\begin{subtable}[c]{\wra\textwidth}
		\resizebox{\textwidth}{!}{\begin{tabular}{|l|lll|lll|}
				\hline
				$\textsf{TPR}$      & \multicolumn{3}{c|}{$v_x$}                                                              & \multicolumn{3}{c|}{$v_y$}                                                             \\ \hline
				$C$ (Truncation level)      & \multicolumn{1}{c|}{-2}            & \multicolumn{1}{c|}{-1}            & \multicolumn{1}{c|}{0} & \multicolumn{1}{c|}{-2}            & \multicolumn{1}{c|}{-1}            & \multicolumn{1}{c|}{0} \\ \hline
				Simulated tempering & \multicolumn{1}{l|}{0.99 (0.07)}   & \multicolumn{1}{l|}{0.99 (0.07)}   & \textbf{0.99} (0.07)   & \multicolumn{1}{l|}{1.00 (0.03)}   & \multicolumn{1}{l|}{0.99 (0.07)}   & 0.98 (0.09)  \\ \hline
				MixedCCA-BIC1       & \multicolumn{1}{l|}{1.00 (0.03)} & \multicolumn{1}{l|}{0.99 (0.07)} & \textbf{0.99} (0.07)  & \multicolumn{1}{l|}{1.00 (0.03)} & \multicolumn{1}{l|}{0.99 (0.07)} & \textbf{0.99} (0.07) \\ \hline
				MixedCCA-BIC2       & \multicolumn{1}{l|}{\textbf{1.00} (0.00)} & \multicolumn{1}{l|}{\textbf{1.00} (0.00) } & \textbf{0.99} (0.07) & \multicolumn{1}{l|}{\textbf{1.00} (0.00)} & \multicolumn{1}{l|}{\textbf{1.00} (0.00)} & \textbf{0.99} (0.07)  \\ \hline
		\end{tabular}}
		\caption{\label{tab:tpr}$\textsf{TPR}$}
		\vspace{5mm}
	\end{subtable}
	
	\begin{subtable}[c]{\wra\textwidth}
		\resizebox{\textwidth}{!}{\begin{tabular}{|l|lll|lll|}
				\hline
				$\textsf{TNR}$      & \multicolumn{3}{c|}{$v_x$}                                                              & \multicolumn{3}{c|}{$v_y$}                                                              \\ \hline
				$C$ (Truncation level)      & \multicolumn{1}{c|}{-2}            & \multicolumn{1}{c|}{-1}            & \multicolumn{1}{c|}{0} & \multicolumn{1}{c|}{-2}            & \multicolumn{1}{c|}{-1}            & \multicolumn{1}{c|}{0} \\ \hline
				Simulated tempering & \multicolumn{1}{l|}{\textbf{1.00} (0.01)}   & \multicolumn{1}{l|}{\textbf{1.00} (0.01)}   & \textbf{0.99} (0.01)  & \multicolumn{1}{l|}{\textbf{1.00} (0.00)}   & \multicolumn{1}{l|}{\textbf{1.00} (0.00)}   & \textbf{0.99} (0.01)  \\ \hline
				MixedCCA-BIC1       & \multicolumn{1}{l|}{0.99 (0.01)} & \multicolumn{1}{l|}{0.99 (0.01)} & 0.98 (0.01) & \multicolumn{1}{l|}{0.98 (0.01)} & \multicolumn{1}{l|}{0.98 (0.01)} & 0.98 (0.01) \\ \hline
				MixedCCA-BIC2       & \multicolumn{1}{l|}{0.98 (0.02)} & \multicolumn{1}{l|}{0.97 (0.02)} & 0.96 (0.05) & \multicolumn{1}{l|}{0.97 (0.02)} & \multicolumn{1}{l|}{0.97 (0.02)} & 0.95 (0.08) \\ \hline
		\end{tabular}}
		\caption{\label{tab:tnr}$\textsf{TNR}$}
	\end{subtable}
	\caption{Mean (and standard deviation) of \textsf{TPR} and \textsf{TNR} of our  method and \textsf{mixedCCA}  for different values of truncation level $C$. }
	\label{tab:2}
\end{table}

\begin{table}[]
	\centering
	\begin{tabular}{|l|ccc|}
		\hline
		Method                 & \multicolumn{3}{c|}{Computation Time (s)}                     \\ \hline
		$C$ (Truncation level) & \multicolumn{1}{c|}{-2}   & \multicolumn{1}{c|}{-1}   & 0     \\ \hline
		Simulated tempering    & \multicolumn{1}{c|}{6.04} & \multicolumn{1}{c|}{7.79} & 16.58 \\ \hline
		MixedCCA-BIC1          & \multicolumn{1}{c|}{1.54} & \multicolumn{1}{c|}{0.81} & 2.28  \\ \hline
		MixedCCA-BIC2          & \multicolumn{1}{c|}{2.15} & \multicolumn{1}{c|}{4.8}  & 6.19  \\ \hline
	\end{tabular}
	\caption{The computation times of all algorithms averaged across 100 continuous datasets for different values of truncation level $C$.}
	\label{tab:mix:cost}
\end{table}

\section{Principal canonical correlation of clinical and proteomic data in Covid-19 patients}\label{sec:covid}
Covid-19 is an infectious disease that is rapidly  sweeping through the world. The disease is  caused by a severe acute respiratory syndrome coronavirus (SARS-CoV-2). There is currently an intense global effort to better understand the virus and find cures and vaccines. We use our methodology to re-analysis a data set produced by \cite{shen:etal:20} that aims to identify biomarkers for early detection of severely ill Covid-19 patients\footnote{For reasons that are still poorly understood, about $80\%$ of patients infected by SARS-CoV-2 experience mild to no symptoms, whereas in about $20\%$ of the cases, patients become severely ill.}. 
To that end, the study enrolled 86 patients (some non-Covid-19 patients, and among the Covid-19 patients, some that developed mild symptoms, and some that became severely ill). The exact protocol for recruiting these patients is unclear. For each patient they measured three (3) physical characteristics (sex, age, and body mass index), twelve (12)  clinical variables as routinely measured from blood samples (white blood cells count, lymphocytes count, C-reactive protein, etc...). Furthermore, the serum of each patient is analyzed by liquid mass spectrometry-based proteomics  to quantify their proteome and metabolome. In \cite{shen:etal:20}, the data is used to build a statistical model to predict whether or not a Covid-19 patient will progress  to a severe state of illness. The dataset of \cite{shen:etal:20} is freely available from the journal website.

We use canonical correlation analysis to re-analyze the data. A common working assumption is that SARS-CoV-2 induces patterns of molecular changes that can be detected in the sera of patients. Canonical correlation analysis may help identify these patterns. To do this we focus on the proteomic data, and we estimate the principal sparse canonical correlation between the physical and clinical variables on one hand and the proteomic variables on the other. See for instance \cite{rousu:etal:13} for a similar analysis on tuberculosis and malaria.

{We pre-process the data by removing all the proteins for which  $50\%$ or more values are missing, leading to a total of $p_y=513$ proteins, and $p_x=16$ clinical and physical variables. The sample size  $n=86$.  Liquid mass spectrometry-based proteomics typically produces a large quantity of missing values (\cite{karpievitch:10,obrien:18}). We make the assumption here that the missing values are driven mainly by detection limit truncation (\cite{karpievitch:10}).  We apply both our algorithm and \textsf{mixedCCA} to this problem, with the same parameter setting as in the simulation test on the mixed datasets (cf.\ Section~\ref{sec:mix:data}). We run  our algorithm for $N=10,000$ iterations. 
	Since we do not know the true canonical pair, we will focus on the estimated canonical correlation to measure the performance of two algorithms. In terms of the estimated canonical correlation, both our algorithm and \textsf{mixedCCA} takes less than 1 second to converge.}

Our estimate of the principal canonical vectors of first dataset ($v_{x\star}$) has only one selected component  (corresponding to C-reactive protein -- CRP) with estimated inclusion probability of $\Pi(\delta_j=1\vert\textbf{Z})= 0.99$. All other physical and clinical variables have inclusion probabilities smaller than $0.1$. We found also that the principal canonical vectors of the proteomic data is also driven by a single protein (P02763, also known as Alpha-1-acid glycoprotein 1 or AGP 1), with estimated inclusion probability of $\Pi(\delta_j=1\vert\textbf{Z})= 0.89$. All other proteins have inclusion probability smaller than $0.1$. Fig.~\ref{fig2} shows the traceplot of  the estimated canonical correlation  $\hat\rho$ between the two data set, as well as the boxplot and autocorrelation function of the MCMC output (after burning in 3/4 of iterations) of the coefficients of CRP and AGP 1 in the quasi-posterior distribution. The fast decay of the autocorrelation functions show a good mixing of the MCMC sampler.

\textsf{MixedCCA} also selects CRP for the clinical dataset and AGP 1 for the proteomic dataset, but both BIC1 and BIC2 criterion select many other variables. $\textsf{mixedCCA-BIC1}$ also selects glucose for clinical dataset and 3 other variables for the proteomic dataset, with estimated canonical correlation $0.90$. $\textsf{mixedCCA-BIC1}$ selects 8 other variables for clinical dataset and 3 additional variables for the proteomic dataset with estimated canonical correlation $0.93$. Although the estimated canonical correlation of \textsf{mixedCCA} is larger than the estimated canonical correlation (0.80) in our algorithm, the highly sparse nature of the estimated canonical vectors estimated from our method is striking. 

Several studies have observed the 
predictive power of C-reative protein (CRP) in the progression of Covid-19 into a severe illness (see for instance \cite{sahu:etal:20} for a meta-analysis). This suggests that the  correlation detected in our analysis between the two datasets is indeed driven by the progression of Covid-19 into a severe illness. Therefore, our analysis suggests that protein AGP 1 may also be playing an important role in the progression of Covid-19 into a severe illness. In Fig. \ref{fig3}, we present the boxplot of CRP and AGP 1 by group of patients. We can see that severe covid patients will have higher value of CRP and AGP 1, compared to non-covid and non-severe patients. We learn from Uniprot\footnote{https://www.uniprot.org}, that AGP 1 functions as transport protein in the blood stream, and appears to function in modulating the activity of the immune system during the acute-phase reaction. Furthermore, AGP 1 appears on the list of differentially expressed proteins in the sera of severely ill Covid-19 patients designed by \cite{shen:etal:20}, and also appeared in the literature as playing a role in the immune system's response to malaria (\cite{friedman:83}). 
\medskip
\begin{center}
	\begin{figure}[h!]
		\centering
		\includegraphics[width=\textwidth]{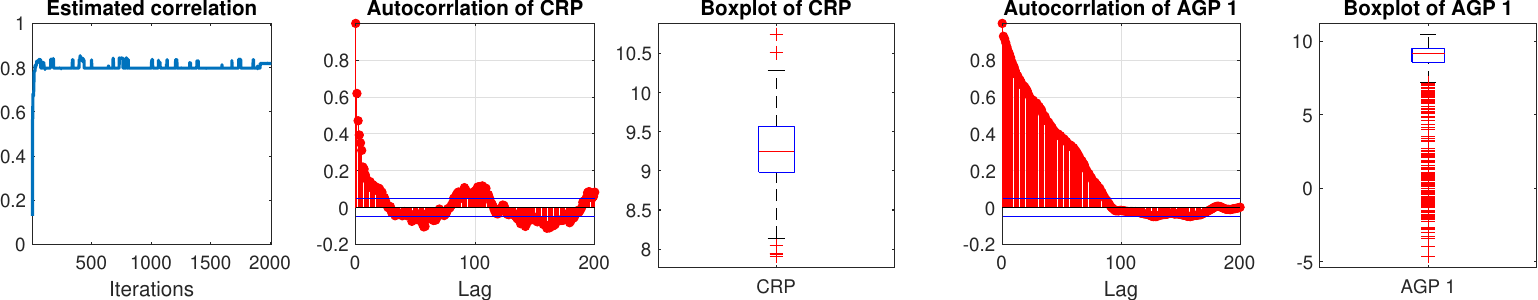}
		\setlength{\floatsep}{-6.0cm}
		\caption{From left to right: The first plot is the trace plot of estimated canonical correlation; The second and third plot is the autocorrelation and boxplot of the coefficient of CRP from MCMC output; The fourth and fifth plot is the autocorrelation and boxplot of the coefficient of AGP 1
			from MCMC output. }
		\label{fig2}
	\end{figure}
	\vspace{-2cm}
\end{center}

\medskip
\begin{figure}
	\centering
	\begin{subfigure}[b]{0.38\textwidth}
		\includegraphics[width=\textwidth]{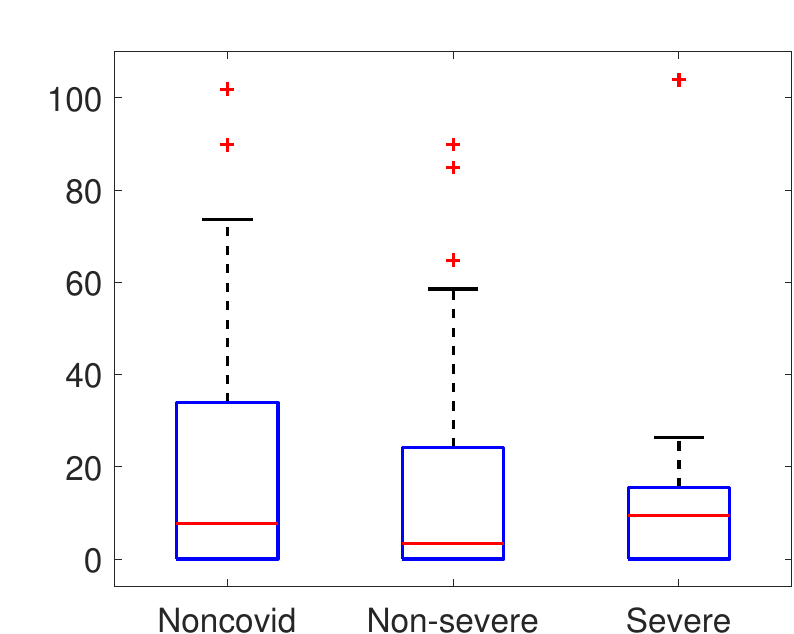}
		\caption{CRP}
	\end{subfigure}\hspace{5mm}
	\begin{subfigure}[b]{0.36\textwidth}
		\includegraphics[width=\textwidth]{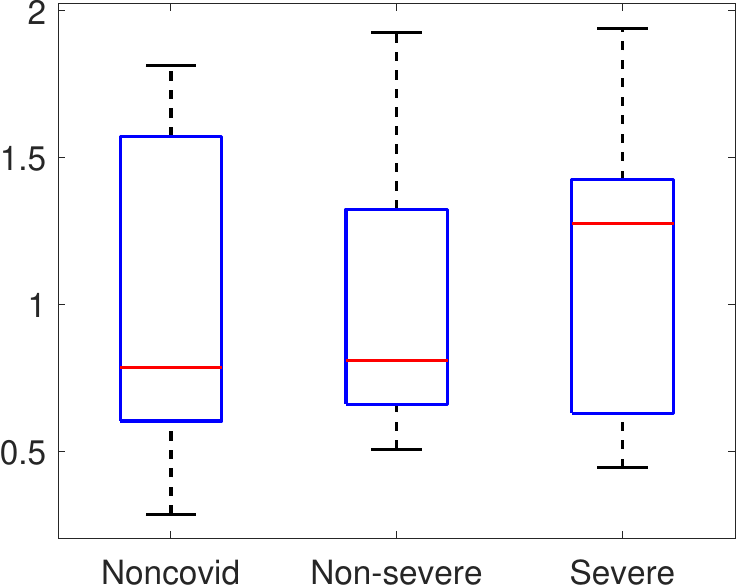}
		\caption{AGP}
	\end{subfigure}
	\caption{Boxplots of (a) CRP and (b) AGP 1 by group of patients.} 
	\label{fig3}
\end{figure}

\section{Conclusion}
\label{sec:conc}

In this work, we have developed a minimax optimal estimation procedure for sparse canonical correlation analysis using a quasi-Bayesian framework.   Our method can be further extended to capture more than one canonical vector, either by deflation, or by reformulating the problem as a higher dimensional canonical correlation analysis estimation problem as in \cite{tan2018sparse}. Furthermore, one can straightforwardly extend our method to solve other generalized eigenvalue problems that arise in other statistical problems, as for instance in Fisher discriminant analysis. At a higher level, the method developed in this work can be viewed as a more statistical implementation of simulated annealing for optimization under sparsity constraints. As such, it can be applied more widely to solve non-convex  optimization problems with sparsity constraints.

\vspace{3cm}

\section{Acknowledgements}
The authors gratefully acknowledge \textit{NSF grant DMS 2015485.} The authors are grateful to Roger Zoh for very helpful discussions.

\bibliographystyle{plain}
\bibliography{Bibliography-MM-MC}

\appendix
\section{Proofs}\label{sec:all:proof}
Throughout the proofs $c_0$ denotes a generic absolute constant that depends only on $\underline{\kappa}$ and $\bar\kappa$ in H\ref{H2}, but whose actual value or expression may change during the text. 
From  the definition of $\Pi(\cdot\vert {\bf Z})$, for any measurable subset $C$ of $\Delta_s\times \rset^p$, by integrating out the non-selected component $\theta-\theta_\delta$, we have
\begin{multline}\label{proof:thm:1:eq1}
	\Pi(C\vert{\bf Z}) = \frac{\sum_{\delta\in\Delta_s}e^{\mathsf{a}\|\delta\|_0}\int_{\rset^p} \textbf{1}_{C}(\delta,\theta)\left(\theta\right)\exp\left(-\frac{\rho_1}{2}\|\theta_\delta\|_2^2 -\frac{\rho_0}{2}\|\theta-\theta_\delta\|_2^2 + \sigma_n \textsf{R}_n(\theta_\delta;{\bf Z})\right)\rmd \theta}{\sum_{\delta\in\Delta_s}e^{\mathsf{a}\|\delta\|_0}\int_{\rset^p} \exp\left(-\frac{\rho_1}{2}\|\theta_\delta\|_2^2 -\frac{\rho_0}{2}\|\theta-\theta_\delta\|_2^2 +\sigma_n\textsf{R}_n(\theta_\delta;{\bf Z})\right)\rmd \theta}\\
	= \frac{\sum_{\delta\in\Delta_s}\left(\frac{1}{p^{\mathsf{u}}}\sqrt{\frac{\rho_1}{2\pi}}\right)^{\|\delta\|_0}\int_{\rset^{\|\delta\|_0}} \textbf{1}_{C}(\delta,(u,0)_\delta)\exp\left(-\frac{\rho_1}{2}\|u\|_2^2 +\sigma_n\bar{\textsf{R}}_n((u,0)_\delta;{\bf Z})\right)\rmd u}{\sum_{\delta\in\Delta_s}\left(\frac{1}{p^{\mathsf{u}}}\sqrt{\frac{\rho_1}{2\pi}}\right)^{\|\delta\|_0}\int_{\rset^{\|\delta\|_0}} \exp\left(-\frac{\rho_1}{2}\|u\|_2^2  +\sigma_n\bar{\textsf{R}}_n((u,0)_\delta;{\bf Z})\right)\rmd u},
\end{multline}
where 
\[\bar{\textsf{R}}_n(\theta;{\bf Z}) \eqdef \textsf{R}_n(\theta;{\bf Z}) - \textsf{R}_n(\theta_\star;{\bf Z}).\]

\subsection{Proof of Theorem \ref{thm:1}}\label{sec:proof:thm:1}
We recall that  $\Delta_s\eqdef \left\{\delta\in\Delta:\;\|\delta\|_0 \leq s\right\}$,  and for $\delta\in\Delta_s$, we let 
\[\cB_{\delta} \eqdef \left\{\theta\in\rset^p:\;\left\|\frac{\theta_\delta\theta_\delta^\T}{\|\theta_\delta\|_2^2} - \theta_\star\theta_\star^\T \right\|_{\textsf{F}}  \leq  M \epsilon \right\},\]
and $\cB_\delta^c$  its complement in $\rset^p$. We then set
\[\mathsf{B} \eqdef \bigcup_{\delta\in\Delta_s}\{\delta\}\times \cB_{\delta},\;\;\mbox{ and }\;\; \mathsf{A} \eqdef \bigcup_{\delta\in\Delta_s}\{\delta\}\times \cB_{\delta}^c.\]  
Clearly we have $\Delta_s\times \rset^p = \cB\cup \mathsf{A}$. Hence our objective is to establish that $\Pi(\mathsf{A}\vert{\bf Z})$ is small.  
We show in Lemma \ref{lem:nc} that the denominator on the right hand side of (\ref{proof:thm:1:eq1}) is bound from below by 
\[\varpi \eqdef e^{-s_\star(\mathsf{u}+1)\log(p)}.\] 
Equation (\ref{proof:thm:1:eq1}) then  implies that
\begin{multline}\label{proof:thm:1:eq2}
	\Pi(\mathsf{A}_1\vert{\bf Z}) \leq \frac{1}{\varpi} \\
	\times \sum_{\delta\in\Delta_s}\left(\frac{1}{p^{\mathsf{u}}}\sqrt{\frac{\rho_1}{2\pi}}\right)^{\|\delta\|_0}\int_{\rset^{\|\delta\|_0}} \textbf{1}_{\cB_\delta^c}((u,0)_\delta)\exp\left(-\frac{\rho_1}{2}\|u\|_2^2 +\sigma_n\bar{\textsf{R}}_n((u,0)_\delta;{\bf Z})\right)\rmd u.
\end{multline}
We show in Lemma \ref{lem:num} that any $\theta\in\rset^p$, such that $\|\theta\|_0\leq s$,
\[
\textsf{R}_n(\theta;{\bf Z}) - \textsf{R}_n(\theta_\star;{\bf Z})  \leq -\frac{\mathsf{gap}}{2}\left(\frac{\underline{\kappa}}{\bar\kappa}\right)^2\|\theta\theta^\T - \theta_\star\theta_\star^\T\|_{\textsf{F}}^2 +c_0 \r_1 \|\theta\theta^\T - \theta_\star\theta_\star^\T\|_{\textsf{F}},\]
for some absolute constant $c_0$ that depends only on $\underline{\kappa}$ and $\bar\kappa$. Therefore, for $\frac{4c_0\r_1}{\textsf{gap}}\left(\frac{\bar\kappa}{\underline{\kappa}}\right)^2 \leq \|\theta\theta^\T - \theta_\star\theta_\star^\T\|_{\textsf{F}}$, we have
\[\textsf{R}_n(\theta;{\bf Z}) - \textsf{R}_n(\theta_\star;{\bf Z}) \leq -\frac{\mathsf{gap}}{4}\left(\frac{\underline{\kappa}}{\bar\kappa}\right)^2\|\theta\theta^\T - \theta_\star\theta_\star^\T\|_{\textsf{F}}^2.\]
Therefore, for $M\geq 4c_0 \left(\frac{\bar\kappa}{\underline{\kappa}}\right)^2$,  (\ref{proof:thm:1:eq2})  becomes 
\begin{multline*}
	\Pi(\mathsf{A}_1\vert{\bf Z}) \leq \frac{1}{\varpi} e^{-\frac{M^2\textsf{gap}}{4}\left(\frac{\underline{\kappa}}{\bar\kappa}\right)^2\sigma_n\epsilon^2} \sum_{\delta\in\Delta_s}\left(\frac{1}{p^{\mathsf{u}}}\sqrt{\frac{\rho_1}{2\pi}}\right)^{\|\delta\|_0}\int_{\rset^{\|\delta\|_0}} \textbf{1}_{\cB_\delta}((u,0)_\delta)\exp\left(-\frac{\rho_1}{2}\|u\|_2^2\right)\rmd u\\
	\leq  \frac{1}{\varpi}  e^{-\frac{M^2\textsf{gap}}{4}\left(\frac{\underline{\kappa}}{\bar\kappa}\right)^2\sigma_n\epsilon^2}\sum_{\delta\in\Delta_s}\left(\frac{1}{p^{\mathsf{u}}}\sqrt{\frac{\rho_1}{2\pi}}\right)^{\|\delta\|_0}\left(2\pi\rho_1^{-1}\right)^{\|\delta\|_0/2} \\
	\leq 2e^{s_\star(\mathsf{u}+1)\log(p)}e^{-\frac{M^2\textsf{gap}}{4}\left(\frac{\underline{\kappa}}{\bar\kappa}\right)^2\sigma_n\epsilon^2} \leq 2e^{-\frac{M^2}{8 \textsf{gap}}\left(\frac{\underline{\kappa}}{\bar\kappa}\right)^2 \sigma_n \r_1^2},
\end{multline*}
under the sample size condition (\ref{cond:ss}), where the third inequality follows from  the assumptions $\mathsf{u}>1$, and $p^{\mathsf{u}-1}>2$.  
This proves the theorem.

\vspace{-0.4cm}
\begin{flushright}
	$\square$
\end{flushright}

We derive here a lower bound on the normalizing constant of the quasi-posterior distribution.

\begin{lemma}\label{lem:nc}
	Suppose that the dataset ${\bf Z}$ satisfies Assumption H\ref{H2}, and $1<\sigma_n\leq p$. Then we can an absolute constant $c_0$ such that $p \geq \max(c_0,e^1 s_\star) $, we have
	\begin{multline}\label{eq:lem:nc}
		\sum_{\delta\in\Delta_s}\left(\frac{1}{p^{\mathsf{u}}}\sqrt{\frac{\rho_1}{2\pi}}\right)^{\|\delta\|_0}\int_{\rset^{\|\delta\|_0}} \exp\left(-\frac{\rho_1}{2}\|u\|_2^2  +\sigma_n\bar{\textsf{R}}_n((u,0)_\delta;{\bf Z})\right)\rmd u \geq e^{-s_\star(\mathsf{u}+1)\log(p))}
	\end{multline}
\end{lemma}
\begin{proof}
	Clearly, the left hand side of (\ref{eq:lem:nc}) is bounded from below by
	\[\left(\frac{1}{p^{\mathsf{u}}}\sqrt{\frac{\rho_1}{2\pi}}\right)^{s_\star}\int_{\rset^{s_\star}} \exp\left(-\frac{\rho_1}{2}\|u\|_2^2  +\sigma_n\bar{\textsf{R}}_n((u,0)_{\delta_\star};{\bf Z})\right)\rmd u.\]
	For any $\theta\in\rset^p$ that has the same support as $\theta_\star$, we have
	\begin{eqnarray*}
		\bar{\textsf{R}}_n(\theta;{\bf Z}) & =  & \frac{\theta^\T\hat A\theta}{\theta^\T\hat B\theta} - \frac{\theta_\star^\T\hat A\theta_\star}{\theta_\star^\T\hat B\theta_\star}  \\
		& =  & \frac{\theta^\T\hat \Sigma\theta}{\theta^\T\hat B\theta} - \frac{\theta_\star^\T\hat \Sigma\theta_\star}{\theta_\star^\T\hat B\theta_\star}  \\
		& = & \frac{\theta^\T\hat \Sigma\theta\left(\theta_\star^\T\hat B\theta_\star - \theta^\T\hat B \theta\right)}{(\theta_\star^\T\hat B\theta_\star)(\theta^\T\hat B \theta)} + \frac{1}{\theta_\star^\T\hat B\theta_\star}\left(\theta^\T\hat \Sigma\theta - \theta_\star^\T\hat \Sigma \theta_\star\right).
	\end{eqnarray*}
	Since $\textsf{R}_n(\cdot;{\bf Z})$ is invariant to rescaling, we can assume without any loss of generality that $\|\theta\|_2=\|\theta_\star\|_2=1$. Therefore for ${\bf Z}$ satisfying H\ref{H2}-(1),we have from Lemma \ref{lem:bound}
	\begin{multline}\label{eq:raigh}
		\left|\bar{\textsf{R}}_n(\theta;{\bf Z}) \right| \leq \left(2\left(\frac{\bar\kappa}{\underline{\kappa}}\right)^2 + 2\left(\frac{\bar\kappa}{\underline{\kappa}}\right)\right) \|\theta\theta^\T - \theta_\star\theta_\star^\T\|_{\textsf{F}} \leq 4\left(\frac{\bar\kappa}{\underline{\kappa}}\right)^2\|\theta\theta^\T - \theta_\star\theta_\star^\T\|_{\textsf{F}}.
	\end{multline}
	It follows from the above observations that for ${\bf Z}$ satisfying H\ref{H2} the left hand side of (\ref{eq:lem:nc})  is  bounded from below by
	\begin{multline*}
		\left(\frac{1}{p^{\mathsf{u}}}\sqrt{\frac{\rho_1}{2\pi}}\right)^{s_\star}\int_{\rset^{s_\star}} \exp\left(-\frac{\rho_1}{2}\|u\|_2^2  -\frac{C}{2} \sigma_n \|uu^\T -\theta_\star\theta_\star^\T\|_{\textsf{F}}\right)\rmd u \\
		\geq \left(\frac{1}{p^{\mathsf{u}}}\sqrt{\frac{\rho_1}{2\pi}}\right)^{s_\star} e^{-C\eta^2 \sigma_n } \int_{\mathcal{S}_0} e^{-\frac{\rho_1}{2}\|u\|_2^2}\rmd u,\end{multline*}
	where $C =8(\bar\kappa/\underline{\kappa})^2$,  $\eta\in (0,1)$ and $\mathcal{S}_0\eqdef\{u\in\rset^{s_\star}:\; \|uu^\T -\theta_\star\theta_\star^\T\|_{\textsf{F}}\leq  2\eta^2\}$. 
	Note that the integral $\int_{\mathcal{S}_0} e^{-\frac{\rho_1}{2}\|u\|_2^2}\rmd u$ is invariant to change of variables by orthogonal matrices. Hence in that integral  we can replace $\theta_\star$ by the unit vector $e=(0,\ldots,0,1)\in\rset^{s_\star}$. Using this and switching to polar coordinates, we write the  integral as 
	\[\int_{\mathcal{S}_0} e^{-\frac{\rho_1}{2}\|u\|_2^2}\rmd u = \int_0^{+\infty} e^{-\frac{\rho_1}{2} r^2} r^{s_\star-1}\rmd r \times \nu\left(\theta\in\mathcal{S}^{s_\star-1}:\; |\sin(\theta)| \leq \eta\right),\]
	where $\nu$ is the surface measure on the unit sphere $\mathcal{S}^{s_\star-1} = \{u\in\rset^{s_\star}:\;\|u\|_2=1\}$, and $\sin(\theta)$ is the sine of the angle between $\theta$ and $e$. The measure $\nu\left(\theta\in\mathcal{S}^{s_\star-1}:\; |\sin(\theta)| \leq \eta\right)$ is equal to twice the spherical cap around the pole $e$ defined by $\eta$. We use the formula of the spherical cap from (\cite{li:11})  to write
	\begin{multline*}
		\nu\left(\theta\in\mathcal{S}^{s_\star-1}:\; |\sin(\theta)| \leq \eta\right) = \frac{4\pi^{\frac{s_\star -1}{2}}}{\Gamma\left(\frac{s_\star -1}{2}\right)} \int_0^{\mathsf{arcsin}(\eta)} \sin^{s_\star -2}(\theta)\rmd \theta \\
		=\frac{4\pi^{\frac{s_\star -1}{2}}}{\Gamma\left(\frac{s_\star -1}{2}\right)} \int_0^{\eta} \frac{x^{s_\star -2}}{\sqrt{1-x^2}} \rmd x \geq \frac{4\pi^{\frac{s_\star -1}{2}}}{\Gamma\left(\frac{s_\star -1}{2}\right)} \frac{\eta^{s_\star-1}}{s_\star -1}.
	\end{multline*}
	Whereas,
	\[\int_0^{+\infty} e^{-\frac{\rho_1}{2} r^2} r^{s_\star-1}\rmd r = \frac{1}{2}\left(\frac{2}{\rho_1}\right)^{\frac{s_\star}{2}} \Gamma\left(\frac{s_\star}{2}\right).\]
	It follows that
	\[\int_{\mathcal{S}_0} e^{-\frac{\rho_1}{2}\|u\|_2^2}\rmd u \geq \frac{2}{s_\star\sqrt{\pi}} \left(\frac{2\pi}{\rho_1}\right)^{\frac{s_\star}{2}} \eta^{s_\star-1}.\]
	We conclude that for ${\bf Z}$ satisfying H\ref{H2}, and any $\eta\in (0,1)$,  the left hand side of (\ref{eq:lem:nc}) is  bounded from below by
	\begin{multline*}
		\frac{2}{\sqrt{\pi}s_\star}\left(\frac{1}{p^{\mathsf{u}}}\right)^{s_\star} e^{-(s_\star-1)\log(1/\eta)} e^{-C\eta^2 \sigma_n} \\
		=\frac{2}{\sqrt{\pi}}\frac{p}{s_\star e^1}\exp\left(-(\mathsf{u}s_\star+1)\log(p)-(s_\star -1)\log(\sqrt{C\sigma_n})\right) \\
		\geq e^{-(\mathsf{u}+1)s_\star \log(p\vee C  \sigma_n)},\end{multline*}
	by taking $\eta = 1/\sqrt{C \sigma_n}$,   and assuming that $p \geq e^1  s_\star $, and $\sqrt{C\sigma_n}\leq p$. 
	This concludes the proof.
\end{proof}

We make use of the following version of the Davis-Kahan $\sin \Theta$ theorem taken from \cite{vu2013} ~ Lemma 4.2. 

\begin{lemma}\label{lem:curvature}
	Let $A$ be a $p\times p$ symmetric semipositive definite matrix and suppose that its eigenvalues satisfies $\lambda_1(A)> \lambda_2(A)\geq \ldots\geq \lambda_p(A)$. If a unit vector $u$  is an eigenvector of $A$ associated to the largest eigenvalue $\lambda_1(A)$, for all $v\in\rset^p$, $\|v\|_2=1$ it holds
	\[\pscal{A}{uu' - vv'} \geq \frac{1}{2}\left(\lambda_1(A) - \lambda_2(A)\right)\|uu'-vv'\|_{\textsf{F}}^2.\]
\end{lemma}

We will need the following technical result.
\begin{lemma}\label{lem:bound}
	For any unit vectors $u,v$ and square matrix $B$ with matching dimensions, we have
	\begin{equation}
		\left|\pscal{B}{uu^\T - vv^\T}\right|  \leq 2\|B\|_{\textsf{op}} \|uu^\T -vv^\T\|_{\textsf{F}},\end{equation}
\end{lemma}
\begin{proof}
	Indeed, we have
	\[\left|\pscal{B}{uu^\T - vv^\T}\right| = \left|(u-v)^\T Bu + v^\T B(u-v)\right| \leq 2\|B\|_{\textsf{op}} \|u-v\|_2.\]
	Similarly, we have $\left|\pscal{B}{uu^\T - vv^\T}\right| 
	\leq  2\|B\|_{\textsf{op}} \|u + v\|_2$. Hence 
	\[\left|\pscal{B}{uu^\T - vv^\T}\right| \leq 2\|B\|_{\textsf{op}} \min\left(\|u-v\|_2, \|u+v\|_2\right).\]
	The result follows by noting that 
	\begin{equation}\label{eq:2F:norm}
		\|uu^\T -vv^\T\|_{\textsf{F}}  \geq \min\left(\|u-v\|_2, \|u+v\|_2\right).\end{equation}
	To see this, note that $\|uu^\T -vv^\T\|_{\textsf{F}} =  \|u-v\|_2 \|u+v\|_2/\sqrt{2} =  \|u-v\|_2 \sqrt{2- \|u-v\|_2^2/2}$. Hence, if $\|u-v\|_2^2 \leq 2$, then we have $\|uu^\T -vv^\T\|_{\textsf{F}} \geq  \|u-v\|_2$. But if $\|u-v\|_2^2> 2$ then $\|uu^\T -vv^\T\|_{\textsf{F}}  > \|u + v\|_2$. Hence the result.
\end{proof}

The next result describes the behavior of the Rayleigh quotient function that yields the posterior contraction result.

\begin{lemma}\label{lem:num}
	Assume H\ref{H2}. For any $\theta\in\rset^p$ such that  $\|\theta\|_0 \leq s$, we have
	\begin{equation}\label{lem:num:eq2}
		\textsf{R}_n(\theta;{\bf Z}) - \textsf{R}_n(\theta_\star;{\bf Z})  \leq-\frac{\mathsf{gap}}{2}\left(\frac{\underline{\kappa}}{\bar\kappa}\right)^2\|\theta\theta^\T - \theta_\star\theta_\star^\T\|_{\textsf{F}}^2 +c_0 \r_1 \|\theta\theta^\T - \theta_\star\theta_\star^\T\|_{\textsf{F}}.\end{equation}
\end{lemma}
\begin{proof}
	Fix $\theta\in\rset^p$ such that $\|\theta\|_0\leq s$. Since the Rayleigh quotient is invariant under rescaling we can assume without any loss of generality that $\|\theta\|_2=1$. We have
	\begin{multline}\label{eq1:proof:contr}
		\bar{\textsf{R}}_n(\theta;{\bf Z}) = \textsf{R}_n(\theta;{\bf Z}) - \textsf{R}_n(\theta_\star;{\bf Z}) = \frac{\theta^\T\hat \Sigma\theta}{\theta^\T\hat B\theta} - \frac{\theta_\star^\T\hat \Sigma\theta_\star}{\theta_\star^\T\hat B\theta_\star} = \frac{\theta^\T \Sigma\theta}{\theta^\T B\theta} - \frac{\theta_\star^\T \Sigma\theta_\star}{\theta_\star^\T B\theta_\star} \\
		+\pscal{\hat\Sigma -\Sigma}{\frac{\theta\theta^\T}{\theta^\T B \theta} - \frac{\theta_\star\theta_\star^\T}{\theta_\star^\T B \theta_\star}} + \pscal{\hat\Sigma}{\left[\frac{\theta\theta^\T}{\theta^\T \hat B\theta} - \frac{\theta\theta^\T}{\theta^\T B\theta}\right] - \left[\frac{\theta_\star\theta_\star^\T}{\theta_\star^\T \hat B\theta_\star} - \frac{\theta_\star\theta_\star^\T}{\theta_\star^\T B\theta_\star}\right]}.
	\end{multline}
	Set $S \eqdef  B^{-1/2}\Sigma B^{-1/2}$,  $w = B^{1/2}\theta/\|B^{1/2}\theta\|_2$, $w_\star = B^{1/2}\theta_\star/\|B^{1/2}\theta_\star\|_2$, and note that $w_\star$ is an eigenvector of $S$ associated to the largest eigenvalue of $S$. Hence by the curvature lemma (Lemma \ref{lem:curvature}) we have
	\[ \frac{\theta^\T \Sigma\theta}{\theta^\T B\theta} - \frac{\theta_\star^\T \Sigma \theta_\star}{\theta_\star^\T B\theta_\star}  = \pscal{S}{ww^\T- w_\star w_\star^\T} \leq -\frac{\mathsf{gap}}{2}\|ww^\T- w_\star w_\star^\T\|^2_{\textsf{F}}.\]
	Let $\I\subseteq \{1,\ldots,p\}$ be the joint  support of $\theta$ and $\theta_\star$ (hence $\|\I\|_0\leq s+s_\star$). Then we can express 
	\[
	\|ww^\T- w_\star w_\star^\T\|_{\textsf{F}} = \left\|(B_{\I,\I})^{1/2}\left(\frac{\theta_{\I}\theta_{\I}^\T}{\theta_{\I}^\T(B_{\I,I})\theta_{\I}} - \frac{\theta_{\star \I}\theta_{\star \I}^\T}{\theta_{\star \I}^\T(B_{\I,\I})\theta_{\star \I}} \right)(B_{\I,\I})^{1/2}\right\|_{\textsf{F}}.\]
	
	We recall that for any square matrix $A$ and invertible matrix $B$, 
	\[\|A\|_{\textsf{F}} = \|B^{-1/2}B^{1/2}AB^{1/2}B^{-1/2}\|_{\textsf{F}}\leq \|B^{-1/2}\|_{\textsf{op}}^2 \|B^{1/2}AB^{1/2}\|_{\textsf{F}}, \]
	where $\|M\|_{\textsf{op}}$ denotes the operator norm of $M$. With these observations in mind, we get
	\begin{multline*}
		\|ww^\T- w_\star w_\star^\T\|_{\textsf{F}} \geq \frac{1}{\|(B_{\I,\I})^{-1/2}\|_{\textsf{op}}^2} \left\|\frac{\theta_{\I}\theta_{\I}^\T}{\theta_{\I}^\T(B_{\I,I})\theta_{\I}} - \frac{\theta_{\star \I}\theta_{\star \I}^\T}{\theta_{\star \I}^\T(B_{\I,\I})\theta_{\star \I}}\right\|_{\textsf{F}}\\
		\geq  \underline{\kappa}\left\|\frac{\theta_{\I}\theta_{\I}^\T}{\theta_{\I}^\T(B_{\I,\I})\theta_{\I}} - \frac{\theta_{\star \I}\theta_{\star \I}^\T}{\theta_{\star \I}^\T(B_{\I,\I})\theta_{\star \I}}\right\|_{\textsf{F}}.
	\end{multline*}
	We note also that for any unit vectors $u,v$ and symmetric invertible matrix $B$ with matching dimension,
	\begin{equation}\label{rk1:diff}
		\left\|\frac{uu^\T}{u^\T Bu} -\frac{vv^\T}{v^\T Bv}\right\|_{\textsf{F}}^2 = \frac{\left(u^\T Bu - v^\T Bv\right)^2}{(u^\T Bu)^2(v^\T Bv)^2}  + \frac{\|uu^\T - vv^\T\|^2_{\textsf{F}}}{(u^\T Bu)(v^\T Bv)} \geq \frac{\|uu^\T - vv^\T\|^2_{\textsf{F}}}{(u^\T Bu)(v^\T Bv)}.\end{equation}
	Hence, under H\ref{H2},
	\[
	\|ww^\T- w_\star w_\star^\T\|_{\textsf{F}}^2 \geq \left( \frac{\underline{\kappa}}{\bar{\kappa}}\right)^2\|\theta_{\I}\theta_{\I}^\T - \theta_{\star \I}\theta_{\star \I}^\T\|_{\textsf{F}}^2 = \left( \frac{\underline{\kappa}}{\bar{\kappa}}\right)^2\|\theta\theta^\T - \theta_{\star}\theta_{\star}^\T\|_{\textsf{F}}^2.
	\]
	In conclusion we have 
	\begin{equation}\label{curvature}
		\frac{\theta^\T \Sigma\theta}{\theta^\T B\theta} - \frac{\theta_\star^\T \Sigma \theta_\star}{\theta_\star^\T B\theta_\star} \leq -\frac{\mathsf{gap}}{2}\left(\frac{\underline{\kappa}}{\bar\kappa}\right)^2\|\theta\theta^\T - \theta_\star\theta_\star^\T\|_{\textsf{F}}^2.
	\end{equation}
	The second term from (\ref{eq1:proof:contr}) can be written as
	\begin{multline*}
		\left|\pscal{\hat\Sigma -\Sigma}{\frac{\theta\theta^\T}{\theta^\T B \theta} - \frac{\theta_\star\theta_\star^\T}{\theta_\star^\T B \theta_\star}} \right| = \left|\pscal{(\hat\Sigma)_{\I,\I} - \Sigma_{\I,\I} }{\frac{\frac{\theta_\I\theta_\I^\T}{\theta^\T B\theta}  - \frac{\theta_{\star \I}\theta_{\star \I}^\T}{\theta_\star^\T B\theta_\star}}{\left\|\frac{\theta_\I\theta_\I^\T}{\theta^\T B\theta}  - \frac{\theta_{\star \I}\theta_{\star \I}^\T}{\theta_\star^\T B\theta_\star}\right\|_{\textsf{F}}}}\right| \left\|\frac{\theta_\I\theta_\I^\T}{\theta^\T B\theta}  - \frac{\theta_{\star \I}\theta_{\star \I}^\T}{\theta_\star^\T B\theta_\star}\right\|_{\textsf{F}}\\
		\leq \max_{M\in\rset^{\I\times \I}:\;\|M\|_{\textsf{F}}=1,\;\textsf{Rank}(M)\leq 2}\;\;\;\left|\pscal{(\hat\Sigma)_{\I,\I} - \Sigma_{\I,\I} }{M}\right|  \times \left\|\frac{\theta_\I\theta_\I^\T}{\theta^\T B\theta}  - \frac{\theta_{\star \I}\theta_{\star \I}^\T}{\theta_\star^\T B\theta_\star}\right\|_{\textsf{F}}.
	\end{multline*}
	And we note from (\ref{rk1:diff}) and Lemma \ref{lem:bound} that for ${\bf Z}$ satisfying  H\ref{H2},  
	\begin{multline}\label{eq:term2:0}
		\left\|\frac{\theta_\I\theta_\I^\T}{\theta^\T B\theta}  - \frac{\theta_{\star \I}\theta_{\star \I}^\T}{\theta_\star^\T B\theta_\star}\right\|_{\textsf{F}}^2 \leq \frac{1}{\underline{\kappa}^4}\pscal{B_{\I\I}}{\theta_\I\theta_\I^T - \theta_{\star,\I} \theta_{\star,\I}^\T}^2 + \frac{1}{\underline{\kappa}^2}\left\|\theta_\I\theta_\I^T - \theta_{\star,\I} \theta_{\star,\I}^\T\right\|_{\textsf{F}}^2\\
		\leq \frac{1}{\underline{\kappa}^2}  \left(1 + 2\left(\frac{\bar\kappa}{\underline{\kappa}}\right)^2\right)   \|\theta\theta^T - \theta_\star \theta_\star^\T\|_{\textsf{F}}^2.\end{multline}
	Therefore for ${\bf Z}$ satisfying  H\ref{H2},
	\begin{equation}\label{eq:term2}
		\left|\pscal{\hat\Sigma -\Sigma}{\frac{\theta\theta^\T}{\theta^\T B \theta} - \frac{\theta_\star\theta_\star^\T}{\theta_\star^\T B \theta_\star}} \right| \leq c_0 \r_1 \|\theta\theta^T - \theta_\star \theta_\star^\T\|_{\textsf{F}}.
	\end{equation}
	
	We process the last term in (\ref{eq1:proof:contr}) as follows.
	\begin{multline}\label{eq:term2:1}
		\pscal{\hat\Sigma}{\left[\frac{\theta\theta^\T}{\theta^\T \hat B\theta} - \frac{\theta\theta^\T}{\theta^\T B\theta}\right] - \left[\frac{\theta_\star\theta_\star^\T}{\theta_\star^\T \hat B\theta_\star} - \frac{\theta_\star\theta_\star^\T}{\theta_\star^\T B\theta_\star}\right]} \\
		= \frac{\theta^\T\hat\Sigma \theta}{\theta^\T\hat B\theta} \pscal{B-\hat B}{\frac{\theta\theta^\T}{\theta^\T B\theta}} - \frac{\theta_\star^\T\hat\Sigma \theta_\star}{\theta_\star^\T\hat B\theta_\star} \pscal{B-\hat B}{\frac{\theta_\star\theta_\star^\T}{\theta_\star^\T B\theta_\star}}\\
		=\left(\frac{\theta^\T\hat\Sigma \theta}{\theta^\T\hat B\theta} - \frac{\theta_\star^\T\hat\Sigma \theta_\star}{\theta_\star^\T\hat B\theta_\star}\right)\pscal{B - \hat B}{\frac{\theta\theta^\T}{\theta^\T B\theta}}
		+ \frac{\theta_\star^\T\hat\Sigma \theta_\star}{\theta_\star^\T\hat B\theta_\star} \pscal{B - \hat B }{\frac{\theta \theta^\T}{\theta^\T B\theta} - \frac{\theta_\star \theta_\star^\T}{\theta_\star^\T B\theta_\star}}.
	\end{multline}
	Hence for  ${\bf Z}$ satisfying  H\ref{H2}, the first term in the last display can be bounded, similar to (\ref{eq:raigh}), as 
	\begin{multline*}
		\left|\left(\frac{\theta^\T\hat\Sigma \theta}{\theta^\T\hat B\theta} - \frac{\theta_\star^\T\hat\Sigma \theta_\star}{\theta_\star^\T\hat B\theta_\star}\right)\pscal{B - \hat B}{\frac{\theta\theta^\T}{\theta^\T B\theta}}\right| \\
		\leq \frac{1}{\underline{\kappa}}\lambda_{\textsf{max}}(\hat B-B,s) \left[4 \left(\frac{\bar\kappa}{\underline{\kappa}}\right)^2\|\theta\theta^\T - \theta_\star\theta_\star^\T\|_{\textsf{F}} \right]\\
		\leq  \frac{4}{\underline{\kappa}}\lambda_{\textsf{max}}(\hat B-B,s) \left(\frac{\bar\kappa}{\underline{\kappa}}\right)^2\|\theta\theta^\T - \theta_\star\theta_\star^\T\|_{\textsf{F}}.
	\end{multline*}
	The rightmost of (\ref{eq:term2:1})  is similar to (\ref{eq:term2}):
	\[\left|\frac{\theta_\star^\T\hat\Sigma \theta_\star}{\theta_\star^\T\hat B\theta_\star} \pscal{B - \hat B }{\frac{\theta \theta^\T}{\theta^\T B\theta} - \frac{\theta_\star \theta_\star^\T}{\theta_\star^\T B\theta_\star}}\right| \leq  c_0 \r_1\|\theta\theta^T - \theta_\star \theta_\star^\T\|_{\textsf{F}}.\]
	In conclusion  the last term in (\ref{eq1:proof:contr}) is bounded from above by
	\begin{equation}\label{eq:term3}
		\left|\pscal{\hat\Sigma}{\left[\frac{\theta\theta^\T}{\theta^\T \hat B\theta} - \frac{\theta\theta^\T}{\theta^\T B\theta}\right] - \left[\frac{\theta_\star\theta_\star^\T}{\theta_\star^\T \hat B\theta_\star} - \frac{\theta_\star\theta_\star^\T}{\theta_\star^\T B\theta_\star}\right]}\right| \leq  c_0 \r_1   \|\theta\theta^T - \theta_\star \theta_\star^\T\|_{\textsf{F}}.
	\end{equation}
	
	We conclude from (\ref{curvature}-\ref{eq:term3}) that for ${\bf Z}$ satisfying  H\ref{H2} 
	\begin{equation*}
		\textsf{R}_n(\theta;{\bf Z})  - \textsf{R}_n(\theta_\star;{\bf Z}) \leq -\frac{\mathsf{gap}}{2}\left(\frac{\underline{\kappa}}{\bar\kappa}\right)^2\|\theta\theta^\T - \theta_\star\theta_\star^\T\|_{\textsf{F}}^2 + c_0 \r_1 \|\theta\theta^T - \theta_\star \theta_\star^\T\|_{\textsf{F}}.
	\end{equation*}
	This ends the proof.
\end{proof}

\section{Proof of Proposition \ref{prop:H2}}\label{sec:proof:prop:H2}
We present the details of this claim for $\hat\Sigma$,  the argument being similar for the other two covariance matrices. For any $J\subset [1:p]$ of size $s$, we have
\[\|\hat\Sigma_{J,J} -\Sigma_{J,J}\|_{\textsf{op}} = \|\Sigma_{J,J}^{1/2}\left(\Sigma_{J,J}^{-1/2}\hat\Sigma_{J,J}\Sigma_{J,J}^{-1/2} -I_s\right)\Sigma_{J,J}^{1/2}\|_{\textsf{op}}\leq \|\Sigma_{J,J}^{1/2}\|^2_{\textsf{op}} \times \|\frac{1}{n}\sum_{i=1}^n U_{i J}U_{i J}^\T -I_s\|_{\textsf{op}},\]
where $U_{i J} \eqdef \Sigma_{J,J}^{-1/2}Z_{i J}$, where $Z_{i J} = (Z_{ij})_{j\in J}$, is mean zero and isotropic. By Theorem 4.6.1 (Equation 4.22) of (\cite{vershynin:18}), provided that $n\geq 4c_0 s\log(p)$ for some absolute constant $c_0>1$, we have
\[\|\hat\Sigma_{J,J} -\Sigma_{J,J}\|_{\textsf{op}} \leq C K^2 \|\Sigma_{J,J}^{1/2}\|^2_{\textsf{op}} \sqrt{\frac{c_0 s\log(p)}{n}},\]
with probability at least $1 -2 p^{-c_0 s}$. Therefore, for any matrix $A\in\rset^{s\times s}$, with $\|A\|_{\textsf{F}}=1$, and $\mathsf{Rank}(A)\leq \alpha$, using the singular value decomposition of $A$, we have
\[\max_{\stackrel{A\in\rset^{s\times s}:\;\|A\|_{\textsf{F}}=1}{ \;\mathsf{Rank}(A)\leq \alpha}} \left|\pscal{\hat\Sigma_{J,J} -\Sigma_{J,J}}{A}\right| \leq \sqrt{\alpha} \|\hat\Sigma_{J,J} -\Sigma_{J,J}\|_{\textsf{op}}\leq  C K^2 \lambda_{\textsf{max}}(\Sigma,s)\sqrt{\frac{c_0\alpha s\log(p)}{n}}.\]
Since the number of subsets of $[1:p]$ of size $s$ is smaller than $p^s$, we conclude with a union bound argument that 
\[\lambda_{\mathsf{max}}^{(\alpha)}(\hat\Sigma - \Sigma,s) \leq C K^2 \lambda_{\textsf{max}}(\Sigma,s) \sqrt{\frac{c_0\alpha s\log(p)}{n}},\]
with probability $1 - 2p^{-(c_0-1)s}$.


\section{MCMC sampling}\label{sec:algo}

\begin{algorithm}[t!]
	\caption{Simulated tempering for sparse canonical correlation analysis}
	\begin{algorithmic}
		\State \textbf{Model Input}: Matrices $\hat A, \hat B$, prior parameters $\rho_0, \rho_1, \mathsf{u}$.
		\State \textbf{MCMC Input}: Number of iterations $N$, batch size $J$, temperatures $1=t_1<\ldots<t_K$, weights $\{c_1,\ldots,c_K\}$, and step-sizes $ \{\eta_1,\ldots,\eta_K\}$.
		\State \textbf{Initialization}: Set the temperature index $k^{(0)} = 1$. Draw $\delta_j^{(0)}\stackrel{i.i.d.}{\sim}\textbf{Ber}(0.5) $ for $j = 1, \ldots, p$,  and independently draw $\theta^{(0)}\sim\textbf{N}(0, I_p)$.  
		
		\For{$t = 0$ to $N-1$, given $(\delta^{(t)}, \theta^{(t)},k^{(t)}) = (\delta,\theta,k)$}{
			\begin{enumerate}
				\item \textbf{Update $\theta$}: 
				Draw the components of $[\bar \theta]_{\bar \delta^c}$independently from $\textbf{N}(0,\rho_0^{-1}t_k)$. Draw $[\bar\theta]_{\bar \delta} \sim P_{k,\delta}([\theta]_{\delta},\cdot)$, where $P_{k,\delta}$ denotes the transition kernel of the MALA with step-size $\eta_k$ and invariant distribution given by  (\ref{cond:dist:eq2}). 
				\vspace{1 ex}
				\item \textbf{Update $\delta$}: 
				Uniformly randomly select a subset $\textsf{J}$ from $\{1,\ldots,p\}$ of size $J$ without replacement, and draw
				$\bar\delta\sim Q_{k,\bar\theta}^{(\mathsf{J})}(\delta,\cdot),$
				where the transition kernel described in (\ref{kernel:gibbs}).
				\vspace{1 ex}
				\item \textbf{Update $k$}:
				Draw $\bar k \sim T_{\bar\delta,\bar\theta}(k,\cdot)$, where $	T_{\delta,\theta}$ is the transition kernel of the Metropolis-Hastings on $\{1,\ldots,K\}$ with invariant distribution given by (\ref{cond:dist:eq3})
				and random walk proposal that has reflection at the boundaries. 
				\vspace{1 ex}
				\item \textbf{New MCMC state}:
				Set $(\delta^{(t+1)},\theta^{(t+1)},k^{(t+1)}) =(\bar \delta, \bar\theta, \bar k)$.
			\end{enumerate}
		}
		\EndFor
		\State \textbf{Output}: $\{(\delta^{(t)},\theta^{(t)}, k^{(t)}): \;0\leq t\leq N \;\;\mbox{ s.t.}\;\;\; k^{(t)} = 1 \}$
	\end{algorithmic}
	\label{algo:1}
\end{algorithm}

We sample from  the simulated tempering distribution (\ref{post:Pi:sCCA:temp}) using  a Metropolis-within-Gibbs strategy. We describe here one iteration of the algorithm, and its transition kernel.  Given $(\delta,\theta,k)$, we perform a three-step update. First, given $k$ and $\delta$, we update $\theta$. We let $[\theta]_\delta$ to denote the $\delta$-selected component of $\theta$ listed in their original order: $[\theta]_\delta \eqdef (\theta_j: j \in \{1 \leq k \leq p: \delta_k = 1\})$, and $[\theta]_{\delta^c} \eqdef (\theta_j: j \in \{1 \leq k \leq p: \delta_k = 0\})$. We  employ the fact that the selected components $[\theta]_\delta$ and the un-selected components $[\theta]_{\delta^c}$ of $\theta$ are independent conditional on $k$ and $\delta$ to update $\theta$. In addition, given $k$ and $\delta$, the components of $[\theta]_{\delta^c}$ are i.i.d. $\textbf{N}(0,t_k\rho_0^{-1})$ and the distribution of  $[\theta]_\delta$ has density on $\rset^{\|\delta\|_0}$ proportional to 
\begin{equation}\label{cond:dist:eq2}
	u\mapsto \exp\left( - \frac{\rho_1}{2t_k}\|u\|_2^2 + \frac{\sigma_n}{t_k}\textsf{R}_n((u,0)_\delta;{\bf Z})\right),\end{equation}
where  the notation $(u,0)_\delta$ denotes the vector in $\rset^p$ such that $[(u,0)_\delta]_\delta = u$.  Hence we update $[\theta]_\delta$ using a  Metropolis adjusted Langevin algorithm (MALA) on $\rset^{\|\delta\|_0}$ with target distribution (\ref{cond:dist:eq2}), and step-size $\eta_k$ (we use different step-sizes for different temperature levels).  Let $M_{\delta,k}$ denote the resulting transition kernel on $\rset^{\|\delta\|_0}$. For more details on the MALA, see e.g., \cite{roberts:tweedie:98}. For convenience, we write $P_{k,\delta}$ to denote the Markov kernel on $\rset^p$ corresponding to the update of $\theta$ just described. Specifically,
\[P_{\delta,k}(\theta,\rmd \theta') = M_{\delta,k}([\theta]_\delta,\rmd [\theta']_\delta) \prod_{j:\;\delta_j=0} \textbf{N}(0,\rho_0^{-1})(\rmd \theta_j'),\]
where $\textbf{N}(\mu,\sigma^2)(\rmd x)$ is a short for the Gaussian measure on $\rset$ with mean $\mu$ and variance $\sigma^2$.

Secondly, we update $\delta$ by applying a Gibbs sampler to the conditional distribution of $\delta$ given $k$ and $\theta$. Note that the conditional distribution of $\delta_j$ given $k,\theta$ and $\delta_{-j}$, where $\delta_{-j} \eqdef (\delta_1,\ldots,\delta_{j-1},\delta_{j+1},\ldots,\delta_p)$, is the Bernoulli distribution $\textbf{Ber}(q_{j})$, with probability of success given by
\begin{equation}\label{cond:dist:eq:1}
	q_{j} \eqdef \left\{1 +\exp\left (-\frac{\mathsf{a}}{t_k}+\frac{1}{2t_k}(\rho_1-\rho_0)\theta_j^2 \right ) \exp\left (\frac{\sigma_n}{t_k}\textsf{R}_n(\theta_{\delta^{(j,0)}};{\bf Z}) - \frac{\sigma_n}{t_k}\textsf{R}_n(\theta_{\delta^{(j,1)}};{\bf Z})\right)\right\}^{-1},\end{equation}
where 
\begin{equation}
	\delta_i^{(j,0)} \eqdef \begin{cases}
		0 & i = j \\ \delta_i & i \neq j
	\end{cases}, \quad \delta_i^{(j,1)} \eqdef \begin{cases}
		1& i = j \\ \delta_i & i \neq j
	\end{cases}.
\end{equation}
Given $k,\theta$ and $j$, let $Q_{k,\theta}^{(j)}$ denote the transition kernel on $\Delta$ which, given $\delta$, leaves $\delta_i$ unchanged for all $i\neq j$, and draws $\delta_j\sim \textbf{Ber}(q_{j})$. We update $\delta$ as follows: randomly draw a subset $\mathsf{J}= \{\mathsf{J}_1,\ldots,\mathsf{J}_J\}$ of size $J$ from $\{1,\ldots,p\}$, and update $\delta$ using the transition kernel on $\Delta$ given by 
\begin{equation}\label{kernel:gibbs}
	Q_{k,\theta}^{(\mathsf{J})}  \eqdef Q_{k,\theta}^{(\mathsf{J}_1)}Q_{k,\theta}^{(\mathsf{J}_2)} \cdots Q_{k,\theta}^{(\mathsf{J}_J)}.\end{equation}
The resulting overall kernel on $\Delta$ is 
\[\bar Q_{k,\theta} = \sum_{\mathsf{J}:\;|\mathsf{J}|=J}{p\choose  J}^{-1}Q_{k,\theta}^{(\mathsf{J})} .\]

Thirdly, given $\delta$ and $\theta$, we update $k$ using a standard Metropolis-Hastings algorithm with a random walk proposal that has reflection at the boundaries. Specifically, at $k$ we propose with equal probability either $k-1$ or $k+1$, except at $1$, where we only propose $2$, and at $K$, where we only propose $K-1$. We write  $T_{\delta,\theta}$ to denote the transition kernel on $\{1,\ldots,K\}$ of this Metropolis-Hastings algorithm with invariant distribution 
\begin{equation}\label{cond:dist:eq3}
	i\mapsto 	\frac{1}{c_i}\exp\left\{\frac{\mathsf{a}}{t_i} \|\delta\|_0-\frac{\rho_1}{2t_i}\|\theta_{\delta}\|_2^2 -\frac{\rho_0}{2t_i}\|\theta-\theta_{\delta}\|_2^2 +  \frac{\sigma_n}{t_i} \textsf{R}_n(\theta_{\delta};{\bf Z})\right\}. \end{equation} 

Lastly, we collect samples by retaining the values of $(\delta,\theta)$ at  iterations at which $k=1$.  In stationarity these samples have distribution  (\ref{post:Pi:sCCA}). 

\subsection{Parameter choices and adaptive tuning}

Throughout the simulation, we specify the parameters of the prior distribution in the following way. We let $\rho_1 = \frac{1}{2}$, and $\rho_0 = n/10$, where $n$ is the sample size, and we  set $\mathsf{u}=1.5$.

Algorithm \ref{algo:1} also depends on the user-defined parameters $J$, $K$, $(t_1,\ldots,t_K)$, $(c_1,\ldots,c_K)$, and $(\eta_1,\ldots,\eta_K)$. The parameter $J$ (the Gibbs sampling batch size) does not greatly impact performance, and setting $J = 100$ works well  in most settings. Efficient selection and tuning of temperatures in simulated tempering has received some attention (\cite{geyer:95, achade11}), and despite some progress (\cite{blazej:etal:13}), to the best of our knowledge, there is no practical and scalable algorithm to do so. In our implementation we use variations of the geometric scaling. We refer the reader to Section \ref{sec:num} for specific choices.

We  tune the step-sizes $\eta=(\eta_1,\ldots,\eta_K)$ of MALA and the weights $(c_1,\ldots,c_K)$ of simulated tempering using adaptive MCMC methods , see e.g., \cite{andrieu:thoms:08}. To tune $\eta_k$, we follow the algorithm proposed in \cite{atchade:rosenthal:05}, with a targeted acceptance probability of $30\%$. For simulated tempering to visit all temperature levels frequently, the weights $(c_1,\ldots,c_K)$ need to be adequately tuned. We refer the reader to \cite{geyer:95} for an extensive discussion of the issue.  This problem can be efficiently solved using the Wang-Landau algorithm for simulated tempering as developed in 
\cite{atchade:liu:10}. We follow this approach here. The fully adaptive MCMC sampler is presented in Algorithm~\ref{algo:1:adapt}.

\begin{algorithm}
	\begin{algorithmic}
		\caption{Adaptive version of simulated tempering for Canonical correlation analysis}
		\label{algo:1:adapt}
		\State \textbf{Model Input}: Matrices $\hat A, \hat B$, prior parameters $\rho_0, \rho_1, \mathsf{u}$.
		\State \textbf{MCMC Input}: Number of iterations $N$, Batch size $J$, temperatures $1=t_1<\ldots,<t_K$.
		\State \textbf{Adaptive MCMC Input}:   $a= 10$ and $ w\in (0,1)$.
		\State \textbf{MCMC Initialization}: Set $k^{(0)} = 1$. Draw $\delta_j^{(0)}\stackrel{i.i.d.}{\sim}\textbf{Ber}(0.5), \forall j = 1, \ldots, p$,  and independently $\theta^{(0)}\sim\textbf{N}(0, I_p)$.  
		\State \textbf{Adaptation Parameters Initialization }: . Set $\ell^{(0)} = \mathbf{0}\in\rset^K$,  $v^{(0)}= (0,\ldots,0)\in\rset^K$, $\nu^{(0)}= (0,\ldots,0)\in\rset^K$, and  choose $c^{(0)}\in (0,\infty)^K$.
		
		\For{$t = 1$ to $N-1$, given $(\delta^{(t)}, \theta^{(t)},k^{(t)}) = (\delta,\theta,k)$, $\ell^{(t)}=\ell$, $c^{(t)} = c$, $v^{(t)}=v$, and $\nu^{(t)}=\nu$}{
			\begin{enumerate}
				\item \textbf{Update $\theta$ and $\ell$}: 
				Draw the components of $[\bar \theta]_{\bar \delta^c}$independently from $\textbf{N}(0,\rho_0^{-1}t_k)$. Draw $[\bar\theta]_{\delta} \sim P_{k,\delta}([\theta]_{\delta},\cdot)$, where $P_{k,\delta}$ denotes the transition kernel of the MALA with step-size $e^{\ell_k}$ and invariant distribution given by (\ref{cond:dist:eq2}). Denote $\alpha$ as the acceptance probability of the MALA update. Set
				\[\bar \ell_k = \ell_k + v_k^{-0.6}(\alpha - 0.3) \;\; \mbox{ and for } i\neq k, \mbox{ set } \bar\ell_i = \ell_i.\]
				\item \textbf{Update $\delta$}: 
				Uniformly randomly select a subset $\textsf{J}$ from $\{1,\ldots,p\}$ of size $J$ without replacement, and draw
				$\bar\delta\sim Q_{k,\bar\theta}^{(\mathsf{J})}(\delta,\cdot),$
				where the transition kernel described in (\ref{kernel:gibbs}).
				\item \textbf{Update $k$, $c$, $v$ and $\nu$}:
				Draw $\bar k \sim T_{\bar\delta,\bar\theta}(k,\cdot)$, where $	T_{\delta,\theta}$ is the transition kernel of the Metropolis-Hastings on $\{1,\ldots,K\}$ with invariant distribution given by (\ref{cond:dist:eq3})
				and random walk proposal with reflection at the boundaries.  We then set 
				\[				\bar c_{\bar k} = c_{\bar k} e^{a},\;\;\;\bar v_{\bar k} = v_{\bar k} +1, \; \bar\nu_{\bar k} = \nu_{\bar k} + 1,\; \mbox{ and for }\; i\neq \bar k, \bar c_i = c_i, \;\bar v_i = v_i,\; \mbox{ and }\; \bar\nu_i =\nu_i.\]	
				\item \textbf{Update $a$ and $\nu$}: If $\|\nu/(\sum_{k=1}^K \nu_k) - 1/K \|_\infty \leq w/K$, then set $a = a/2, \nu =  \mathbf{0}\in\rset^K$.
				\item \textbf{New MCMC state}:
				Set $(\delta^{(t+1)},\theta^{(t+1)},k^{(t+1)}) =(\bar \delta, \bar\theta, \bar k)$, $\ell^{(t+1)} = \bar\ell$, $ c^{(t+1)} = \bar c$,  $v^{(t+1)} = \bar v$, and $\nu^{(t+1)} = \bar\nu$.
			\end{enumerate}
		}
		\EndFor
		\State \textbf{Output}: $\{(\delta^{(t)},\theta^{(t)}, k^{(t)}): \;0\leq t\leq N \;\;\mbox{ s.t.}\;\;\; k^{(t)} = 1 \}$
	\end{algorithmic}
	
\end{algorithm}

\section{Coupled Markov chains for mixing time estimation}\label{sec:mixing:est}

\label{sec:umcmc}
At least empirically, simulated tempering is well-known to improve mixing when dealing with multimodal distributions (\cite{geyer:95,liu2008monte}). 
However, rigorous results are far less well-established. Using a Markov kernel decomposition approach, (\cite{woodard:etal:09}) gives a lower bound on the spectral gap of simulated tempering in terms of the spectral gaps of the component kernels and the so-called projection kernel. However, applying  their result to a specific problem remains non-trivial. Furthermore, their lower bound decays exponentially fast in the number of components in the partition, which clearly limits its relevance in our setting. Using a similar Markov kernel decomposition technique, (\cite{ge:etal:18}) has a more explicit upper bound on the mixing time of simulated tempering. However their result applies to a different algorithm than the one considered here, and they consider a specific form of the target distribution that does not include (\ref{post:Pi:sCCA:temp}). 

Given the lack of theoretical mixing time analysis of simulated tempering, we take a more empirical approach based on the unbiased Markov Chain Monte Carlo framework in (\cite{biswas2019estimating,jacob:umcmc}). Let $\{X^{(t)},\;t\geq 0\}$ be the Markov chain generated by our simulated tempering algorithm, where $X^{(t)} = (\delta^{(t)},\theta^{(t)},k^{(t)}) \in \mathsf{X}$. Let $P$ denote its transition kernel (which is described in Section~\ref{sec:mixing:est} in the supplementary material). Following \cite{jacob:umcmc}, we construct a coupling $\check P$ of $P$ with itself: that is, a transition kernel on $\mathsf{X}\times \mathsf{X}$ such that $\check P((x,y), A\times \mathsf{X}) = P(x,A)$, $\check P((x,y),\mathsf{X}\times B) = P(y,B)$, for all $x,y\in\mathsf{X}$, and all measurable sets $A,B$. Furthermore, $\check P((x,x), \mathcal{D}) =1$ where  $\mathcal{D} \eqdef\{(x,x):\;x\in\mathsf{X}\}$. The construction of the Markov kernel $\check P$ is described in Section \ref{sec:mixing:est} in the supplementary material. 

Given $\check P$, a lag $L\geq 1$, and an initial distribution as given in the initialization step of Algorithm~\ref{algo:1}, we  simulate a bivariate Markov chain $\{(X_t, Y_{t-L}),\;t\geq L\}$ as follows. First draw $X^{(0)}\sim\Pi^{(0)}$ and  $Y^{(0)}\sim\Pi^{(0)}$. Next, for $1\leq t\leq L$, we draw $X_t\vert (X_0,Y_0,X_1,X_2,\ldots,X_{t-1}) \sim P(X_{t-1},\cdot)$. Then  for $t> L$, we draw 
\[(X_{t},Y_{t-L})\vert \left\{(X_{t-1},Y_{t-L-1}),\ldots,(X_L,Y_0),X_{L-1},\ldots,X_0\right\} \sim \check P\left((X_{t-1},Y_{t-L-1}),\cdot\right).\]
In other words, at each time $t> L$ we attempt to couple the two chains while maintaining the correct marginals. We define $\tau^{(L)} \eqdef\inf\left\{t\geq L:\; X_{t} = Y_{t-L}\right\}$, and have the following:

\begin{proposition}\label{prop:mixing}
	Let $\{X^{(t)},\;t\geq 0\}$ be the Markov chain generated by the simulated tempering algorithm, 
	and let $\bar\Pi^{(t)}$ denote the distribution of $X^{(t)}$. For all $t\geq 0$, we have
	\begin{equation}\label{eq:coupl}
		\|\bar\Pi^{(t)} - \bar\Pi\|_\tv \leq \PE\left[\max\left(0,\left\lceil \frac{\tau^{(L)} - L-t}{L}\right\rceil\right)\right].\end{equation}
\end{proposition}
\begin{proof}
	See Section \ref{proof:prop:mixing}.
\end{proof}

This inequality implies that by simulating multiple copies of the bivariate chain, and approximating the expectation in~\eqref{eq:coupl} by Monte Carlo, we can actually estimate the mixing time of our algorithm. This gives us the possibility to investigate empirically the mixing time of our sampler with some theoretical guarantees. 

\subsection{Coupled Markov Chains}

We describe here the specific coupled Markov chain employed to estimate the mixing time plots presented in Section \ref{sec:mixing}.
We refer the reader to \cite{biswas2019estimating} and \cite{jacob:umcmc} for more details on the construction of such coupled kernels. We modify  Algorithm~\ref{algo:1} to  construct the coupled kernel $\check P$. It suffices here to describe one iteration of the coupled chain. At some iteration $ t \geq 1$, suppose that $(\delta^{(1, L + t)},\theta^{(1, L+t)},k^{(1, L+t)}) = (\delta^{(1)},\theta^{(1)},k^{(1)})$ and $(\delta^{(2,t)},\theta^{(2,t)},k^{(2,t)}) = (\delta^{(2)},\theta^{(2)},k^{(2)})$. 

In step 1, to update $\theta^{(1)}$ and $\theta^{(2)}$, we partition the indices $\{1, \ldots,p\}$ into four groups: $G_{ab}=\{j:\; \delta^{(1)}_{j}=a, \delta^{(2)}_{j}=b\}$ for $a, b = 0, 1$. To update the components of $\theta^{(1)}_{G_{00}}$ and $\theta^{(2)}_{G_{00}}$, for any $j \in G_{00}$ we first draw a common standard normal random variables $Z_j$, and then obtain $\bar\theta^{(i)}_j = t_{k^{(i)}}\rho_0^{-1} Z_j $ for $i = 1,2$. To update the components of $\theta^{(1)}_{G_{01}}$ and $\theta^{(2)}_{G_{01}}$, for any $j \in G_{01}$ we again first draw a common standard normal random variables $Z_j$, and then obtain $\bar\theta^{(1)}_j = t_{k^{(1)}}\rho_0^{-1} Z_j $, and simultaneously draw $\bar\theta^{(2)}_j$ using MALA with proposal $\theta^{(2)}_j + \eta_{k^{(2)}} \nabla \log \pi(\theta^{(2)}_j)  + \sqrt{2 \eta_{k^{(2)}} } Z_j $ , where $\pi(\theta^{(2)}_j)$ is the marginal posterior distribution of $\theta^{(2)}_j$. Notice that the joint distribution of $[\theta^{(2)}]_{\delta^{(2)}}$ is given by $W_{k^{(2)},\delta^{(2)}}$, whose density is proportional to (\ref{cond:dist:eq2}). A similar update procedure is used for updating the components of $\theta^{(1)}_{G_{10}}$ and $\theta^{(2)}_{G_{10}}$. To update the components of $\theta^{(1)}_{G_{11}}$ and $\theta^{(2)}_{G_{11}}$, we draw reflection-coupled MALA proposals in \cite{biswas2019estimating}, and then for the acceptation step, $\theta^{(1)}_{G_{11}}$ and $\theta^{(2)}_{G_{11}}$ share the same uniform random variables.

In step 2, to update $\delta^{(1)}$ and $\delta^{(2)}$, we first make use of the same randomly drawn subset $\mathsf{J}$. For $i = 1,2$, drawing $\bar \delta^{(i)} \sim Q_{k, \theta}^{(\mathsf{J})} (\delta^{(i)}, \cdot)$ is equivalent to let $ \bar \delta^{(i)}_{-\mathsf{J}} = \delta^{(i)}_{-\mathsf{J}}$, and for any $j\in\mathsf{J}$, draw  $\bar \delta^{(i)}_j\sim \textbf{Ber}(q^{(i)}_{j})$ which we implement in the following way. We first draw a common uniform number $u_j\sim \textbf{Uniform}(0,1)$, then we obtain  $\bar \delta^{(i)}_j = \mathbf{1}\{q^{(i)}_{j} \leq u_j\}$ for $i = 1,2$.

In step 3, to update $k^{(1)}$ and $k^{(2)}$, we use a common uniform random number to make the proposal move, and a common uniform random number for the acceptation step.

\begin{remark}
	Note that although the empirical mixing time estimation method of \cite{biswas2019estimating}  described above only applies to Markov chains {with fixed parameters}, we have applied it here to Algorithm~\ref{algo:1:adapt}, which is an MCMC sampler with adaptively tuned parameters.  We conjecture that the  unbiased MCMC methodology remains approximately valid for well-constructed adaptive MCMC samplers. However the question deserves more research.
\end{remark}

\subsection{Proof of Proposition \ref{prop:mixing}}\label{proof:prop:mixing}
Using the notations established in Section \ref{sec:algo} the transition kernel of the simulated tempering algorithm on $\Xset\eqdef \Delta\times\rset^p\times   \{1,\ldots,K\}$ is given by
\[P((\delta,\theta,k);(\delta',\rmd\theta',k')) = P_{\delta,k}(\theta,\rmd \theta') \left\{\sum_{\mathsf{J}:\;|\mathsf{J}|=J}\;{p\choose J}^{-1} Q^{(\mathsf{J})}_{k,\theta'}(\delta,\delta')\right\}T_{\delta',\theta'}(k,k'),\]
and we call $\bar P$ the transition kernel of the coupled chain on $\Xset\times \Xset$ as described in Section \ref{sec:mixing:est}. The kernel $P$ is a standard Metropolis-within-Gibbs kernel to sample from the density (\ref{post:Pi:sCCA:temp}) that is positive everywhere. Therefore, $P$ is phi-irreducible, aperiodic and has invariant distribution $\bar\Pi(\cdot\vert {\bf Z})$ by construction. Furthermore, for any nonempty compact set $\Cset$ of $\rset^p$, the set $\bar \Cset \eqdef \Cset\times\Delta\times \{1,\ldots,K\}$ is a small set for $P$, and it is easy to see from the construction of the coupled chain that, 
\[\min_{x,y\in\bar\Cset}\;\bar P^{n_0}\left((x,y);\mathcal{D}\right) >0,\;\;\mbox{ with } n_0 = \max\left(K,\frac{p}{J}\right).\]
Therefore, according to Proposition 4 of (\cite{jacob:umcmc}) to establish the finiteness of the average meeting time $\PE(\tau^{(L)})$, it suffices to show that there exist a drift function $V:\;\Xset\to [1,\infty)$,  $\lambda \in (0,1)$, and $b<\infty$ such that 
\begin{equation}\label{drift:to:show}
	P V(x) \leq \lambda V(x) + b\textbf{1}_{\bar\Cset}(x),\;\;\mbox{ for all }\;\; x=(\delta,\theta,k)\in\Xset,\end{equation}
for some small $\bar\Cset$ of the form $\bar\Cset_L\eqdef \{x\in\Xset:\; V(x) \leq L\}$. We show (\ref{drift:to:show})  in three steps, with
\[V(\delta,\theta,k) \eqdef 1 + \frac{1}{t_k} \|\theta_\delta\|_2^2.\]

\paragraph{\textsf{Step 1: Action of the kernel $T_{\delta,\theta}$}}\;\; We first show that for all $(\theta,\delta,k)\in\Xset$,
\begin{equation}\label{drift:T}
	T_{\delta,\theta} V(\delta,\theta,k) \leq V(\delta,\theta,k) + c_0,\end{equation}
for some  constant $c_0$.  To show this, we find it easy to reason in a slightly more general terms. Consider a discrete distribution on $\{1,\ldots,K\}$, given by
\[\pi(k) \propto \frac{1}{c_k}e^{-U/t_k},\]
for some increasing sequence $\{c_k,\;1\leq k\leq K\}$, and for some nonnegative constant  $U$.  Consider a Metropolis-Hastings algorithm to sample from  $\pi$ with a proposal $q$ on $\{1,\ldots,K\}$ such that at $j$, we propose to move only to $j-1$ or $j+1$ for equal probability (at $1$ we propose to move only to $2$, and at $K$ we propose to move only to $K-1$). 
Call $M$ the transition kernel of that Metropolis-Hastings, and for some nonnegative constant $B$, define
\[V(j) =  \frac{B}{t_j},\;\;j=1,\ldots,K.\]
By the definition of the Metropolis-Hasting kernel, we have
\begin{multline*}
	M V(j) = V(j) + \sum_{j'=1}^K \left(V(j')-V(j)\right)\min\left(1,\frac{\pi(j')q(j',j)}{\pi(j)q(j,j')}\right)q(j,j') \\
	= V(j) +\sum_{j'=1}^KR(j,j')q(j,j'),\end{multline*}
where
\[R(j,j') = \left(V(j')-V(j)\right)\min\left(1,\frac{\pi(j')q(j',j)}{\pi(j)q(j,j')}\right).\]
Note that $V(j+1)\leq V(j)$, and therefore $R(j,j+1)\leq 0$.  Whereas
\begin{multline*}
	R(j,j-1) = B\left(\frac{1}{t_{j-1}} - \frac{1}{t_j}\right)\min\left(1,\frac{c_j}{c_{j-1}}\frac{q(j-1,j)}{q(j,j-1)}e^{-\left(\frac{1}{t_{j-1}}-\frac{1}{t_j}\right)U}\right)\\
	\leq \left(\frac{B}{U}\right) \frac{2c_j}{c_{j-1}} \left(\frac{1}{t_{j-1}} - \frac{1}{t_j}\right)Ue^{-\left(\frac{1}{t_{j-1}}-\frac{1}{t_j}\right)U}\\
	\leq 2e^{-1} \left(\max_j\frac{c_j}{c_{j-1}}\right) \frac{B}{U} =C,
\end{multline*}
where we use the fact that $q(j',j)/q(j,j')\leq 2$, and the observation  that $te^{-t}\leq e^{-1} $ for all $t\geq 0$. Using these we have
\[MV(1) = V(1) + R(1,2) \leq V(1).\]
\[MV(K) = V(K) + R(K,K-1)\leq V(K) +  C.\]
For $2\leq j\leq K-1$,
\[MV(j) = V(j) + \frac{1}{2}R(j,j-1) + \frac{1}{2}R(j,j+1) \leq V(j) + \frac{C}{2}.\]
Hence, for all $1\leq j\leq K$, it holds
\[MV(j)  \leq V(j) + 2e^{-1} \left(\max_j\frac{c_j}{c_{j-1}}\right) \frac{B}{U}.\]
We can apply this result to the kernel $T_{\delta,\theta}$ with $U =\frac{\rho_1}{2}\|\theta_\delta\|_2^2 + \frac{\rho_0}{2}\|\theta-\theta_\delta\|_2^2 - \sigma_n\textsf{R}_n(\theta_\delta;{\bf Z}) -\textsf{a}\|\delta\|_0$, and $B = \|\theta_\delta\|_2^2$.  Under assumption H\ref{H1}, $\textsf{R}_n(\theta_\delta;{\bf Z})\leq 1$. Hence for $\rho_1\|\theta_\delta\|_2^2 \geq 4(\sigma_n + \mathsf{a}p)$, the chosen $U$ is non-negative and we get
\begin{multline*}
	T_{\delta,\theta} V(\delta,\theta,k) \leq  V(\delta,\theta,k) \\
	+ 2e^{-1} \left(\max_j\frac{c_j}{c_{j-1}}\right) \frac{\|\theta_\delta\|_2^2}{\frac{\rho_1}{2}\|\theta_\delta\|_2^2 + \frac{\rho_0}{2}\|\theta-\theta_\delta\|_2^2 - \sigma_n\textsf{R}_n(\theta_\delta;{\bf Z}) -\textsf{a}\|\delta\|_0} \\
	\leq V(\delta,\theta,k) + \frac{8e^{-1}}{\rho_1} \left(\max_j\frac{c_j}{c_{j-1}}\right),\end{multline*}
for $\|\theta_\delta\|_2 \geq L$, for $L$ taken large enough.

\paragraph{\textsf{Step 2: Accounting for the kernel $\bar Q$}}\;\;
For consistency in the notation, we write summations as integrals with respect to the counting measure. Using (\ref{drift:T}) and the definition of $\bar Q$, we have for all $(\delta,\theta,k)\in\Xset$ such that $\|\theta_\delta\|_2\geq L$ for some appropriately large $L$,
\begin{multline*}
	\int_\Delta \bar Q_{\theta,k}(\delta,\rmd \delta')\int T_{\theta,\delta'}(k,\rmd k')V(\theta,\delta',k') \leq \int_\Delta \bar Q_{\theta,k}(\delta,\rmd \delta')V(\theta,\delta',k) + c_0\\
	= \sum_{\mathsf{J}:\;|\mathsf{J}|=J}\;{p\choose J}^{-1} \int_\Delta  Q^{(\mathsf{J})}_{\theta,k}(\delta,\rmd \delta')V(\theta,\delta',k) + c_0.
\end{multline*}
Given a selection $\mathsf{J}=\{j_1,\ldots,j_J\}\subseteq \{1,\ldots,p\}$, and $j_i\in\mathsf{J}$, we  have
\begin{multline*}
	\int_\Delta \tilde Q_{k,\theta}^{(j_i)}(\delta,\rmd\delta')V(\theta,\delta',k)  = V(\theta,\delta,k)  + \left(q_{j_i}  - \delta_{j_i}\right)\frac{\theta_{j_i}^2}{t_k} \leq V(\theta,\delta,k)  +\left(1 - \delta_{j_i}\right)\frac{\theta_{j_i}^2}{t_k},
\end{multline*}
where $q_j$ is as given in (\ref{cond:dist:eq:1}). It follows that
\[
\int_\Delta  Q^{(\mathsf{J})}_{\theta,k}(\delta,\rmd \delta')V(\theta,\delta',k) \leq 
V(\theta,\delta,k) +\frac{1}{t_k}\sum_{i=1}^J  \theta_{j_i}^2\textbf{1}_{\{\delta_{j_i} = 0\}}.\]
We conclude that for $\|\theta_\delta\|_2\geq L$,
\begin{multline}\label{control:TQJ}
	\int_\Delta \bar Q_{\theta,k}(\delta,\rmd \delta')\int T_{\theta,\delta'}(k,\rmd k')V(\theta,\delta',k') \leq 
	V(\theta,\delta,k) \\
	+\frac{1}{t_k}\sum_{\mathsf{J}:\;|\mathsf{J}|=J}\;{p\choose J}^{-1}\left\{\sum_{i=1}^J  \theta_{\mathsf{J}_i}^2\textbf{1}_{\{\delta_{\mathsf{J}_i} = 0\}}\right\}  + c_0.\end{multline}

\paragraph{\textsf{Step 3: Drift condition for $P$}}\;\; We recall that under the kernel $P_{\delta,k}$ the components $\{\theta_j',\;j:\; \delta_j=0\}$ are drawn independently from the Gaussian distribution $\textbf{N}(0,\rho_0^{-1})$. Therefore, for $\|\theta_\delta\|_2\geq L$, using (\ref{control:TQJ}), we have
\begin{multline}\label{prop:mcmc:step3:eq1}
	\int_\Xset P\left((\theta,\delta,k);(\rmd\theta',\rmd\delta',\rmd k')\right) V(\theta',\delta',k') \\
	\leq \int_{\rset^p} P_{\delta,k}(\theta,\rmd \theta')\left[V(\theta',\delta,k) +\frac{1}{t_k}\sum_{\mathsf{J}:\;|\mathsf{J}|=J}\;{p\choose J}^{-1}\left\{\sum_{i=1}^J  (\theta_{\mathsf{J}_i}')^2\textbf{1}_{\{\delta_{\mathsf{J}_i} = 0\}}\right\}  + c_0 \right]\\
	\leq  1 + \frac{1}{t_k}\int_{\rset^{\|\delta\|_0}} M_{k,\delta}(\theta_\delta,\rmd v)\|v\|_2^2 + \frac{J}{\rho_0 t_k} + c_0,\end{multline}
where $M_{k,\delta}$ is the kernel of the MALA with target distribution proportional to
\[u\mapsto \exp\left( - \frac{\rho_1}{2t_k}\|u\|_2^2 + \frac{\sigma_n}{t_k}\textsf{R}_n((u,0)_\delta;{\bf Z})\right).\]
It remains to deal with the term $\int_{\rset^{\|\delta\|_0}} M_{k,\delta}(\theta_\delta,\rmd v)\|v\|_2^2$. For clarity sake let's work is a slightly more general setting. Suppose that we have a density on $\rset^d$, $d\geq 1$,  that is proportional to $e^{-m(u)}$ for some function $m$ of the form
\[m(u) \eqdef  \frac{\rho}{2}\|u\|_2^2 + \ell(u),\]
for some bounded function $\ell$.  Let $q_\eta(u,\cdot)$ be the density of the proposal distribution $\textbf{N}\left(\left(1 - \frac{\rho\eta^2}{2}\right)u,\eta^2 I_{d}\right)$, and define $\textsf{R}(u)\eqdef\{v\in\rset^d:\;\alpha(u,v) < 1\}$, where 
\[\alpha(u,v) = \min\left(1,\frac{e^{-m(v)}q_\eta(v,u)}{e^{-m(u)}q_\eta(u,v)}\right).\]
Let $L$ denote the resulting transition kernel on $\rset^p$. 
We have
\begin{eqnarray*}\label{PV}
	\int_{\rset^{d}} L(u,\rmd v)\|v\|_2^2 &=& \|u\|_2^2 + \int \alpha(x,y)\left(\|v\|_2^2 -\|u\|_2^2\right)q_\eta(u,v)\rmd v\nonumber\\
	&=& \|u\|_2^2 + \int_{\textsf{R}(u)} \left[\frac{e^{-m(v)}q_\eta(v,u)}{e^{-m(u)}q_\eta(u,v)} -1 \right]\left(\|v\|_2^2 -\|u\|_2^2\right)q_\eta(u,v)\rmd v \nonumber\\
	&&+ \int_{\rset^d}\left(\|v\|_2^2 -\|u\|_2^2\right)q_\eta(u,v)\rmd v.\end{eqnarray*}
We can write
\[\|v\|_2^2 -\|u\|_2^2 =2\pscal{u}{v-u} +\|v-u\|_2^2.\]
Integrating both sides,  we get
\begin{multline}\label{bound1}
	\int_{\rset^d}\left(\|v\|_2^2 -\|u\|_2^2\right)q_\eta(u,v)\rmd v = -\rho\eta^2\|u\|_2^2 + \frac{\rho^2\eta^4}{4}\|u\|_2^2 + \eta^2 d\leq -\frac{3}{4}\rho\eta^2\|u\|_2^2  +\eta^2d,
\end{multline}
by choosing $\eta$ such that $\eta^2\rho \leq 1/2$. We also have
\begin{multline*}
	\frac{e^{-m(v)}q_\eta(v,u)}{e^{-m(u)}q_\eta(u,v)}=\exp\left(m(u) -m(v) -\frac{1}{2\eta^2}\left\|u-v+\frac{\rho\eta^2}{2}v\right\|^2 +\frac{1}{2\eta^2}\left\|v-u+\frac{\rho\eta^2}{2} u\right\|^2\right).
\end{multline*}
If $v\in\textsf{R}(u)$, we necessarily have $\frac{e^{-m(v)}q_\eta(v,u)}{e^{-m(u)}q_\eta(u,v)}<1$, which translates to:
\[m(v) - m(u)> -\frac{1}{2\eta^2}\left\|u-v+\frac{\rho\eta^2}{2}v\right\|^2 +\frac{1}{2\eta^2}\left\|v-u+\frac{\rho\eta^2}{2}u\right\|^2.\]
Noting that $m(u) = (\rho/2)\|u\|_2^2 + \ell(u)$, where $\ell$ is bounded by $b_0$, we infer that for $v\in\textsf{R}(u)$,
\begin{multline*}
	\frac{\rho}{2}\left(\|u\|_2^2 - \|v\|_2^2\right)\leq 2 b_0 +\frac{1}{2\eta^2}\left\|u-v+\frac{\rho\eta^2}{2}v\right\|^2 -\frac{1}{2\eta^2}\left\|v-u+\frac{\rho\eta^2}{2}u\right\|^2\\
	=2b_0 + \frac{\rho^2\eta^2}{8}\left(\|v\|_2^2 - \|u\|_2^2\right) -\frac{\rho}{2}\pscal{v-u}{v+u}\\
	= 2b_0 + \frac{\rho^2\eta^2}{8}\left(\|v\|_2^2 - \|u\|_2^2\right) +\frac{\rho}{2}\left(\|u\|_2^2 - \|v\|_2^2\right).
\end{multline*}
Hence, for $v\in\textsf{R}(u)$,
\[\|u\|_2^2 - \|v\|_2^2 \leq \frac{16 b_0}{\rho^2\eta^2},\]
which we use to write
\begin{multline}\label{bound2}
	\int_{\textsf{R}(u)} \left[\frac{e^{-m(v)}q_\eta(v,u)}{e^{-m(u)}q_\eta(u,v)} -1 \right]\left(\|v\|_2^2 -\|u\|_2^2\right)q_\eta(u,v)\rmd v \\
	= \int_{\textsf{R}(u)} \left[1 - \frac{e^{-m(v)}q_\eta(v,u)}{e^{-m(u)}q_\eta(u,v)} \right]\left(\|u\|_2^2 - \|v\|_2^2\right)q_\eta(u,v)\rmd v\ \leq  \frac{16 b_0}{\rho^2\eta^2}.
\end{multline}
We combine (\ref{PV})-(\ref{bound2}) to conclude that 
\[\int_{\rset^{d}} L(u,\rmd v)\|v\|_2^2 \leq \|u\|_2^2 -\frac{3}{4}\rho\eta^2\|u\|_2^2  +\eta^2d + \frac{16 b_0}{\rho^2\eta^2}.\]
Hence we can find $b_1$ (for instance $b_1 = 4\rho^{-1}(d + 16b_0\rho^{-2}\eta^{-4})$) such that for $\|u\|_2^2>b_1$, it holds
\[\int_{\rset^{d}} L(u,\rmd v)\|v\|_2^2 \leq \left(1 - \frac{\rho\eta^2}{2}\right)\|u\|_2^2.\]

This results applied to $M_{k,\delta}$ together with (\ref{prop:mcmc:step3:eq1}) implies that there exist $\lambda\in (0,1)$ (for instance $\lambda =\frac{\rho\eta^2}{2t_K}$), and  $L<\infty$ such that
\[PV(\delta,\theta,k) \leq \lambda V(\delta,\theta,k),\;\;\mbox{ for all} (\delta,\theta,k)\notin \bar \Cset_L.\]
With similar (but simpler) calculations we check that
\[\sup_{(\delta,\theta,k)\in \bar \Cset_L}\; PV(\delta,\theta,k) \leq b,\]
for some finite constant $b$. This establishes the drift condition
\[PV(\delta,\theta,k) \leq \lambda V(\delta,\theta,k) + b\textbf{1}_{\bar\Cset_L}(\delta,\theta,k),\;\;\mbox{ for all } (\delta,\theta,k)\in \Xset.\]
Hence the result.

\subsection{Empirical studies of our algorithm}
\subsubsection{On sparse CCA computational barrier} \label{sec:samplesize}
It was conjectured by (\cite{gao:etal:17}) that it is not possible to solve the sparse CCA problem in polynomial time at the statistical rate $\epsilon$ obtained in (\ref{rate:scov}), in the data regime $n = o(s_\star^2\log(p))$. The authors made a compelling argument for this conjecture by showing that any such estimator for the sparse CCA  can be used to solve the planted clique problem in a regime where it is widely believed to be computationally intractable.  Since our estimator achieves the rate $\epsilon$ under the weaker condition $n\geq C_0 s_\star\log(p)$, we have the opportunity to test empirically this conjecture.  

In our simulation, we let $\Sigma_{x}\in\mathbb{R}^{(p/2)\times(p/2)}$ and $\Sigma_{y}\in\mathbb{R}^{(p/2)\times(p/2)}$ 
share the same structure, namely, a block diagonal matrix with five blocks, each of dimension $p/10 \times p/10$, where the $(j,j')$-th element of each block takes value $0.8^{|j-j'|}$. We let $\lambda_1=0.9, (v_{x\star})_j =(v_{y\star})_j= 1/\sqrt{3}$ for $j \in \{ 1, 6, 11\}$, and $(v_{x\star})_j = (v_{y\star})_j = 0$ otherwise. Therefore, the true density level  is $s_\star = 6$. For each $p \in \{500, 2000, 5000\}$, we generate data  from the model described in Section \ref{sec:data:model}  with two values of the sample size $n$, namely $\lceil s_\star^{1.5} \log(p)\rceil$ and $\lceil s_\star^{2.5} \log(p)\rceil$. We use the sample covariance matrices as estimators of $\Sigma_x$, $\Sigma_y$ and $\Sigma_{xy}$, and set the scaling parameter $\sigma_n = 2n$ to construct the extended posterior distribution   $\bar\Pi$ in (\ref{post:Pi:sCCA:temp}).  We sample from $\bar\Pi$ using Algorithm~\ref{algo:1:adapt}, with the set of temperatures $\{1, 1/0.9, 1/0.8, 1/0.7\}$. Since in this particular data model, the largest value of the (population) Rayleigh quotient is $\lambda_1=0.9$, proximity of the sample Rayleigh quotient $\textsf{R}_n(\cdot;{\bf Z})$ to $\lambda_1$ along the MCMC iterations is a good empirical measure of  mixing.  

\begin{figure}
	\centering
	\begin{subfigure}[b]{0.32\textwidth}
		\includegraphics[width=\textwidth]{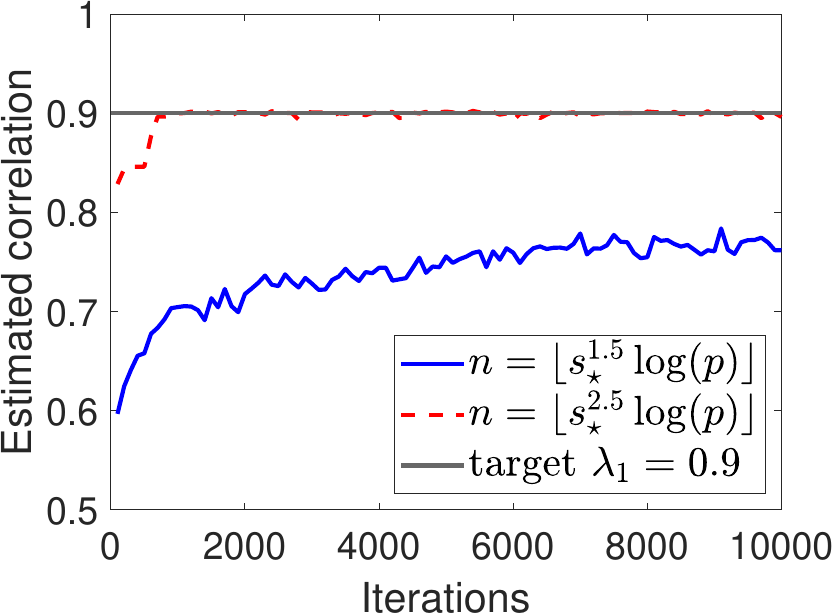}
		\caption{$p=500$}
	\end{subfigure}\hfill
	\begin{subfigure}[b]{0.32\textwidth}
		\includegraphics[width=\textwidth]{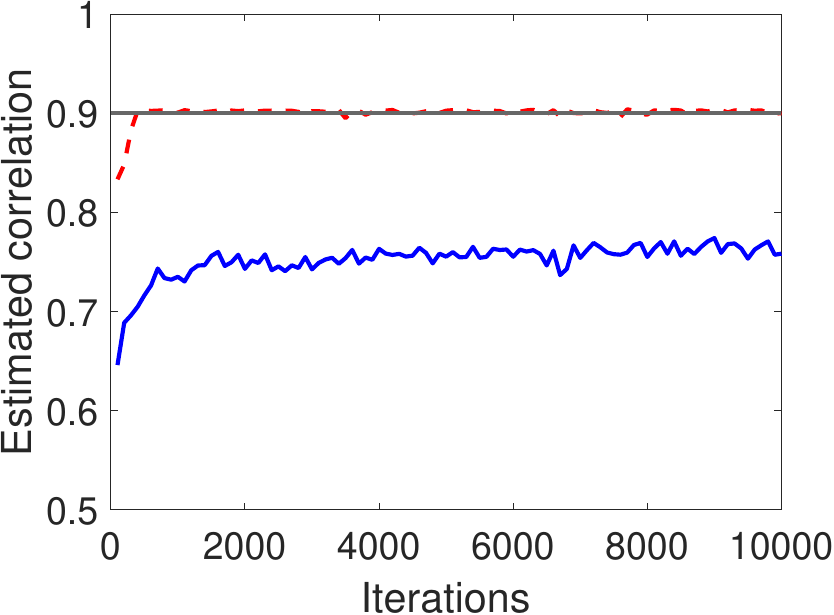}
		\caption{$p=2000$}
	\end{subfigure}\hfill
	\begin{subfigure}[b]{0.32\textwidth}
		\includegraphics[width=\textwidth]{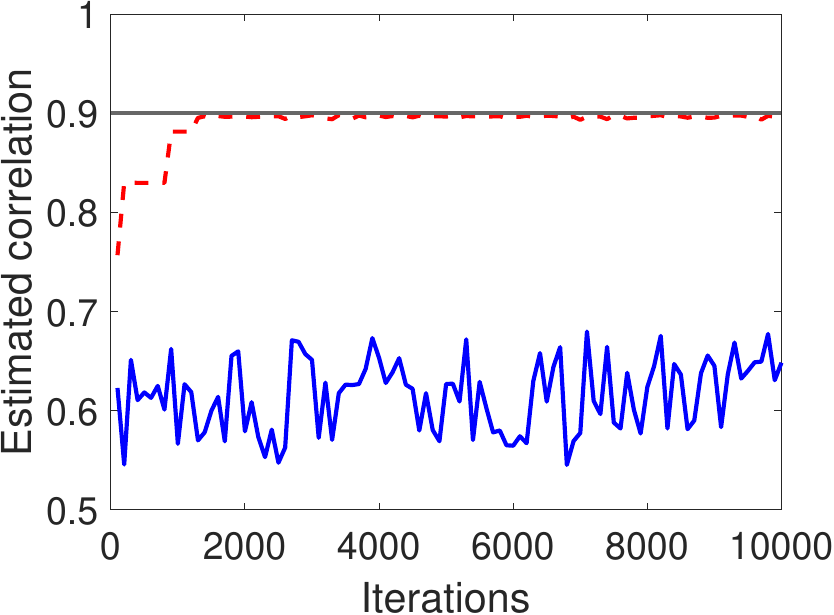}
		\caption{$p=5000$}
	\end{subfigure}
	\caption{Estimated canonical correlation along the MCMC iterations, averaged over 30 data repetitions for different values of dimension $p$ and sample size $n$. } 
	\label{sample_size}
\end{figure}

We run each MCMC sampler for $N = 10,000$ iterations, repeated $30$ times (each time with a newly generated dataset). At each iteration time, we average the values of $\textsf{R}_n(\cdot;{\bf Z})$ across the $30$ repetitions.  
Fig.~\ref{sample_size} shows the plot of the averaged sample Rayleigh quotient along iterations. 
The difference in behavior is striking. We clearly see that for all values of $p$, the sample Rayleigh quotient $\textsf{R}_n(\cdot;{\bf Z})$ corresponding to $n = \lceil s_\star^{2.5} \log(p)\rceil$ quickly converges to the population Rayleigh quotient $\lambda_1=0.9$, whereas the one corresponding to $n = \lceil s_\star^{1.5} \log(p)\rceil$ fails to converge even after 10,000 iterations. 
This suggests that the condition $n\geq C_0 s_\star^2\log(p)$ is indeed  needed for the simulated tempering sampler to mix well, which appears to confirm the conjecture by (\cite{gao:etal:17}).

\medskip

\medskip

\subsubsection{Empirical mixing time of Algorithm \ref{algo:1}}\label{sec:mixing}
We  investigate more carefully 
the mixing time of Algorithm \ref{algo:1} as a function of the dimension $p$, using  the coupled chain approach of (\cite{biswas2019estimating,jacob:umcmc}) as described in Section \ref{sec:umcmc}. We focus on a data-rich setting where the sample size $n=p/2$. 
%
Now, let us describe the implementation details. We let  $\Sigma_{x}$, $\Sigma_{y}$, $v_{x\star}$ and $v_{y\star}$ all have the same structures as in Section~\ref{sec:samplesize} and set $\lambda_1=0.9$. We generate datasets from the model in Section \ref{sec:data:model} for each $p \in \{100, 200, \ldots, 5000\}$, with sample size  $n = p/2$. The extended posterior distribution   $\bar\Pi$ in (\ref{post:Pi:sCCA:temp}) is constructed in the same way as in Section~\ref{sec:samplesize}, except with the set of temperatures $\{1, 1/0.9, 1/0.8, 1/0.7, 1/0.6\}$. We set the lag $L=p$ and the maximum iterations $N=10 p + 1000$. 
For each value of $p$, we repeat the simulation $50$ times to estimate the distribution of the meeting time $\tau^{(L)}$ of the chain. More precisely, using $\varepsilon=0.1$, we estimate the mixing time of the chain as the first iteration $t$ for which the Monte Carlo estimate of the right hand side of (\ref{eq:coupl}) is less than $\varepsilon$. Fig.~\ref{fig1}  below shows the plot of the mean of meeting times and the estimated mixing times as functions of $p$. The results suggest that Algorithm \ref{algo:1:adapt} has a mixing time that scales roughly linearly in the dimension $p$. 

\begin{rem}
	As far as we know, the existing literature on simulated tempering gives only general guidelines on choosing the temperatures (\cite{geyer:95, achade11}). The implementation of these guidelines remains challenging, and typically requires further adaptive MCMC methods (\cite{blazej:etal:13}). In our case, the Rayleigh quotient responds very well to temperature tuning, and in particular does not require high temperatures to mix well. As a result, we have chosen to maintain some very simple temperature scaling, and these work very well in the our experiments.
\end{rem}

\medskip
\begin{center}
	\begin{figure}[h!]\centering
		\includegraphics[width=.8\textwidth]{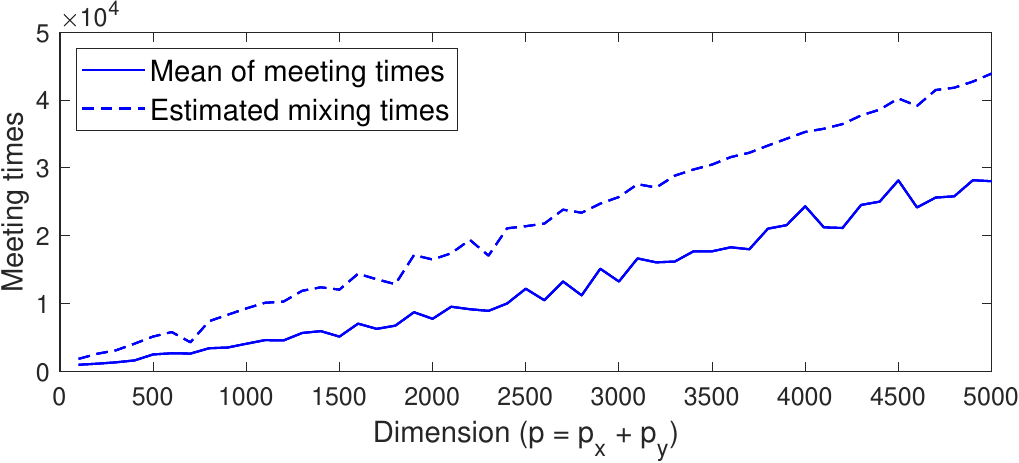}
		\setlength{\floatsep}{-6.0cm}
		\caption{The mean of meeting times versus the estimated mixing times. The estimated mixing times are with respect to the total variation distance 0.1. }
		\label{fig1}
	\end{figure}
	
\end{center}

\medskip

%
%

\end{document}